\documentclass[article]{jss-tr}

\usepackage{thumbpdf,lmodern}
\usepackage{framed}
\usepackage{amsmath}
\usepackage[utf8]{inputenc}
\usepackage{longtable}
\usepackage{booktabs}
\usepackage[ruled,vlined,linesnumbered]{algorithm2e}
\usepackage[dvipsnames]{xcolor}
\usepackage{listings}
\usepackage{subcaption}
\usepackage{url}
\usepackage{amssymb}

\newcommand{\IGNORE}[1]{}

\newcommand{\Diis}{Departamento de Informática e Ingeniería de Sistemas} 
\newcommand{\Unizar}{Universidad de Zaragoza, Spain} 
 
\newcommand{\Hib}{Hiberus, Spain}

\newcommand{\rea}{\mathrm{I\!R}}
\newcommand{\nat}{\mathrm{I\!N}}
\definecolor{vlgray}{gray}{0.9}

\graphicspath{{./}}
 
\author{Simona Bernardi, José Merseguer\\\Unizar
   		\And  Raúl Javierre\\\Hib
	}

\title{\pkg{tegdet}: An extensible \proglang{Python} Library for Anomaly Detection using Time-Evolving Graphs}
\Plaintitle{\pkg{tegdet}: A  Python Library for Anomaly Detection using Time-Evolving Graphs}
\Shorttitle{\pkg{tegdet}: A Python Library for TEG-based anomaly detection}

\Abstract{
This paper presents a new \proglang{Python} library for anomaly detection in unsupervised learning approaches. The input for the library is a univariate time series representing observations of a given phenomenon. Then, it can identify \emph{anomalous epochs}, i.e., time intervals where the observations are above a given percentile of a baseline distribution, defined by a dissimilarity metric.
Using \emph{time-evolving graphs} for the anomaly detection, the library leverages valuable information given by the inter-dependencies among data. 
Currently, the library implements $28$ different dissimilarity metrics, and it has been designed to be easily extended with new ones. Through an API, the library exposes a complete functionality to carry out the anomaly detection. Summarizing,  to the best of our knowledge, this library is the only one publicly available, that based on dynamic graphs, can be extended with other state-of-the-art anomaly detection techniques. 
Our experimentation shows promising results regarding the execution times of the algorithms and the accuracy of the implemented techniques.
Additionally, the paper provides guidelines for setting the parameters of the detectors to improve their performance and prediction accuracy.
}

\Address{
  Simona Bernardi, José Merseguer\\
  \Diis, \Unizar\\
  E-mail: \email{\{simonab,jmerse\}@unizar.es}
  \emph{and}\\
  Raúl Javierre,\\
  \Hib\\
  E-mail: \email{rjavierre@hiberus.com}
}

\begin{document}


\section{Introduction} \label{sec:intro}

Anomaly detection is a research field initiated in the $20$th century by the statistics community, being the work of~\cite{Grubbs69}, among others, a reference. The computer science community started to research and apply anomaly detection techniques in the $80$s of the previous century. These techniques are currently applied in many different and heterogeneous fields, such as network intrusion detection, malware detection, fraud detection, data center monitoring, social networks or social media research. Undoubtedly, for an industrial adoption of these techniques, also for researchers to leverage them, it is mandatory the existence of software packages, whose implementation favors ease of use. Moreover, 
considering the myriad of existing anomaly detection techniques, software extensibility is another important property. In this context, extensibility refers to the capability of easily adding new anomaly detection techniques to the software package, and not needing to fully learn the package implementation.

Since many are the fields where spotting anomalies is a need, then many are the anomaly detection techniques. Some of them can identify outliers in unstructured collections of multi-dimensional data points. However, in many domains, such as finance, security, communication media or health care, data
exhibit inter-dependencies, which provide valuable information to detect outlier data. Hence, graphs have been prevalently used to represent structural/relational information among data in many domains, e.g.~\cite{Aggarwal11,Wu18}, then raising the \emph{graph anomaly detection problem}.~\cite{ATK15} identified four reasons that highlight the need of graphs for anomaly detection: a) {inter-dependent nature of the data}, b) {powerful representation} for such inter-dependencies, c) {relational nature of problem domains} and d) {robust machinery}, for graphs to serve as adversarial tools also. More specifically, \emph{dynamic} or \emph{time-evolving graphs}, i.e., sequences of static graphs, have been extensively used in the literature for detecting outliers in time series. The book of~\cite{Gupta14} surveys many anomaly detection techniques in this field, while~\cite{Febrinanto22} expose current challenges in the dynamic evolution of graph data.

This paper aligns with definitions given by~\cite{ATK15}, which are based on the outlier definition given by~\cite{Hawkins}: ``an observation that
differs so much from other observations as to arouse suspicion that it was generated by a different mechanism''. Then, the {\em dynamic graph anomaly detection problem} is defined as ``{given} a sequence of (plain/attributed) graphs, {find} (i) the timestamps that correspond to a {change} or {event}, as well as (ii) the top-k nodes, edges, or parts of the graphs that contribute most to the change''. Currently, our work deals with the first part of the definition, while the finding of ``top parts of the graphs'' will be the subject of future versions.

Regarding publicly available implementations of dynamic graph anomaly detection techniques, the recent survey by~\cite{MWXYZSXA21} identifies eight works and their corresponding repositories. 
In the following, we review these works.
SedanSpot~\citep{sedanspot} implements in \proglang{C++}~\citep{stroustrup} an algorithm to detect anomalies from an edge stream in near real-time, where anomalies are edges connecting sparsely-connected parts of the graph (bridge edges) and possible occurring during intense bursts of activity.
AnomalyDAE~\citep{AnomalyDAE} is a \proglang{Python}~\citep{van2007python} code implementing a deep learning anomaly detector, that aims at finding nodes whose patterns deviate significantly from the majority of reference nodes.
ChangeDAR~\citep{Changedar}, also developed in \proglang{Python}, detects change points using sensors placed on nodes or edges. It can be applied to detecting electrical failures, traffic accidents or other events in road traffic graphs.
AnomRank~\citep{ANOMRANK} implements a \proglang{C++} algorithm that detects sudden weight changes along an edge and also sudden 
structural changes to the graph. 
A GitHub repository\footnote{https://github.com/danieltan07/dagmm} implements DAGMM~\cite{DAGMM}, as a set of \proglang{Python} scripts.
DAGMM is an unsupervised anomaly detection technique that uses a deep autoencoder to generate a low-dimensional representation and
reconstruction error for each input data point, which is fed into a Gaussian Mixture Model. Then, it optimizes the parameters and the model jointly.
F-FADE~\citep{F-FADE} is an anomaly detection technique, implemented in \proglang{Python}, that applies to edge streams. It develops a technique, that requires constant memory, to efficiently model time-evolving distributions between node-pairs. 
MIDAS~\citep{MIDAS} is an online technique focussed on detecting microcluster anomalies, groups of suspicious similar edges, instead of individually edges. It has been implemented in \proglang{C++}.
DeepSphere~\citep{DeepSphere} develops and implements in \proglang{Python} an unsupervised algorithm to address case-level and nested level anomaly detection. It does not require outlier-free data as input and can reconstruct normal patterns.
Besides these eight works, we have also identified many other interesting works, such as~\cite{SAPE,TADDY,Addgraph}. However, we have not found available implementations of these works, which is the subject of this paper.

This paper presents a novel \proglang{Python} library for anomaly detection, based on dynamic graphs.
The input of the approach, also of the library, must be a univariate time series representing observations of a given phenomenon. The output identifies {\em anomalous epochs}, i.e.,  time intervals where the observations are above a given percentile of a distribution, which is defined by a given dissimilarity metric. Currently, the library implements $28$ different dissimilarity metrics, and it has been designed to be easily extended. 
Therefore, a prominent feature of the library is the ease for users to introduce new anomaly detection techniques.

Unlike our work, most of the previously reviewed software presents the implementation of a graph-based anomaly detection technique, which has been usually developed by the very same authors, to solve a specific domain problem.
These implementations aim at proving the efficiency and correctness of the technique. Thus, none of these works are specifically interested on defining a \proglang{Python} library exposing a proper API, as a mean to offer the functionalities of the anomaly detection for being integrated into other packages. 
Another salient difference is that there is no concern in these works for developing a software design that 
supports the extensibility of the package, with new anomaly detection techniques proposed by other authors. 
On the contrary, our library is completely focussed on being a reference, in \proglang{Python}, for integrating implementations of state-of-the-art anomaly detection techniques, based on dynamic graphs. 
Summarizing, our library is ready to be used by other software implementations and ready to be extended with other techniques.  
The only conditions for a technique, such as those developed in the reviewed works, to be integrated in the library are that: 
a) it accepts, as input, a univariate time series and b) it fits in an unsupervised learning approach, as described in Section~\ref{sec:models}. 

Finally, our experimentation shows promising results regarding the execution times of the algorithms (i.e., time to build the prediction model
and time to get the {\em anomalous epochs}) and the accuracy of the implemented anomaly detectors for the considered dataset.
Additionally, the paper offers guidelines, for setting the parameters of the anomaly detectors, which improve
the performance and prediction accuracy.

The rest of the paper is structured as follows.
Section~\ref{sec:models} presents the anomaly detection approach implemented by the library.
Section~\ref{sec:impl} overviews the software package and the use of the library. 
Section~\ref{sec:application} recalls an experiment to detail the use of the library.
Section~\ref{ss:basic_assessment} assesses the quality of the library.
Section~\ref{sec:guidelines_predictions} offers guidelines for the users to set the parameters of the library, so to get the most of it. 
Section~\ref{sec:summary} concludes the work.
Appendices~\ref{app:technical},~\ref{app:extensibility}~and~\ref{app:repo} offer necessary technical details for the correct use and extension of the library. 


\section{Models and algorithms} \label{sec:models}

\begin{figure*}[hbt]
	\includegraphics[width=\textwidth]{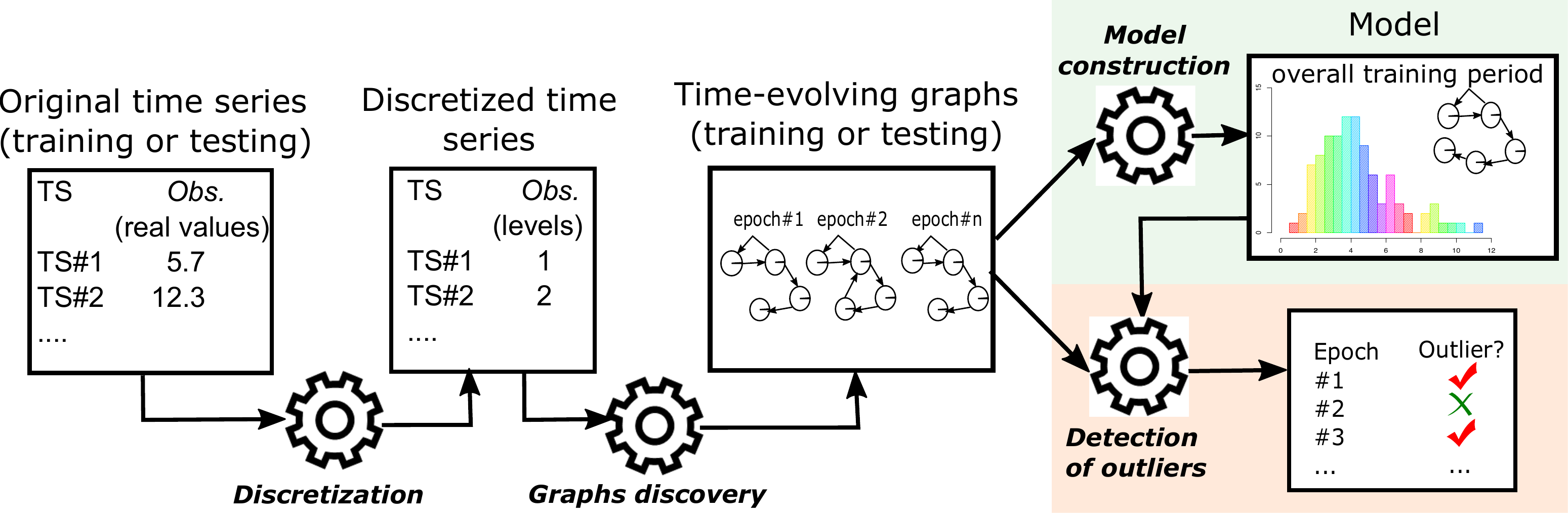}
	\caption{Unsupervised learning approach}
	\label{fig:approach}			
\end{figure*}

Time Evolving Graphs (TEGs,~\cite*{SS2011}) are sequences of static graphs. TEGs have been successfully used for {\em anomaly detection} in unsupervised learning approaches~\citep{ATK15}. Figure~\ref{fig:approach} summarizes the unsupervised learning approach followed in this work. In such approaches, a {\em prediction model} is built first, from \emph{training datasets}, which represent the behaviour under study. 
Then, this model is fed with \emph{testing datasets} to detect outliers in the future behaviour. In our approach, input datasets, {\em training} and {\em testing}, are univariate time series, and they are used to create TEGs. The prediction model is obtained from the {\em training} TEGs, while the outliers detection uses the prediction model and the {\em testing} TEGs.

The creation of TEGs is a two-step process: discretization and graphs discovery. First, the original time series ({\em training} or {\em testing} dataset) is discretized, which means to map real valued observations onto an ordered set of \emph{levels}. Second, the discretized time series is used to create (discover) the TEGs. Concretely, by partitioning the overall time period of the series into equal sized consecutive time intervals, namely \emph{epochs}. 
Therefore, each graph synthesizes observations of the original time series during an epoch.
Hence, the granularity of the TEGs depends on two factors: the number of epochs and the {\em level} of discretization. 
The former defines the number of graphs, which is equal to the number of epochs.
The latter affects the size of the graphs, in terms of the number of nodes and edges.

In a subsequent step, a {\em prediction model} is built from the {\em training} TEGs. The model is made of a \emph{global graph} and a \emph{baseline distribution}. The former is a graph synthesizing the TEGs, and therefore, the observations from the entire training dataset. 
The latter is the distribution of the dissimilarities between each graph of the {\em training} TEGs and the global graph. Since different dissimilarity metrics can be used to calculate such distribution, then, there are as many different anomaly detectors as dissimilarity metrics.

The {\em anomaly detection} encompasses two steps: the generation of {\em testing} TEGs and the {\em detection of outliers}. 
The latter computes the dissimilarities between each graph of the  {\em testing} TEGs and the global graph of the prediction model. 
An epoch is then labeled as anomalous if the dissimilarity between its corresponding graph and the global graph is above 
the $k^{th}$ percentile of the baseline distribution, where $k$ is a parameter of the detector.

In the following, the four steps in Figure~\ref{fig:approach} are formalized: discretization, graphs discovery, model construction and detection of outliers.

\subsection{Discretization}\label{SS:discretization}
A univariate time series, of a real valued variable, represents an ordered set of observations (or data points, $o$) taken at regular time instants (timestamps, $ts$), e.g., every 30 minutes. Discretization means mapping real-valued observations into an ordered set of \emph{levels}, that is, integer values, through a discretization function. Hence, the function depends on the observations and on the chosen number of \emph{levels}, the level of discretization.
The discretization function is defined prior to construct the prediction model, i.e., considering only the {\em training} dataset.

\paragraph{Discretization function.}
Let   $\mathcal{T}=\{ \langle ts_k,o_k\rangle \}_{k\in K\subset\nat}$ be a univariate time series. We denote as: 
\[ m= \mathop{min}_{k\in K}\ \{ o_k \}, \ M = \mathop{max}_{k\in K}\ \{ o_k\} \]
the minimum and maximum data points, respectively.

The set of data points is partitioned into $n$ intervals,  {\em level of discretization}, of equal length 
$[m,M] = \mathop{\cup_{i=0}^{n-1}} [x_i,x_{i+1})$. Hence, the real value domain is partitioned into $n+2$ intervals:
\[
\rea =  (\infty,m)  \mathop{\cup_{i=0}^{n-1}} [x_i,x_{i+1}) \cup [M,\infty)
\]
where $x_0=m$ and $x_{n}=M$.

Then, the discretization function ${\mathcal D}: \rea \rightarrow \nat$  is defined as follows:
\begin{equation}\label{e:discretization}
{\mathcal D}(x)  = 
	\begin{cases} 
        i,  \mathrm{\ if\ } x\in [x_i,x_{i+1}),\  i=0,\dots,n-1\\
        n,  \mathrm{\ if\ } x\in (\infty,m)\\
        n+1,  \mathrm{\ if\ } x\in [M,\infty)
	\end{cases}
\end{equation}

Observe that, $n$ will not be assigned to observations in the training dataset. However, it is possible that it could be assigned to some observations in the testing dataset, during the detection phase.

\subsection{Graphs discovery}\label{SS:discovery}
Graph discovery is the second, and last, step in the generation of TEGs.
This step actually produces a sequence of graphs $\mathcal{G} = \{ G_j \}_{j\in J\subset\nat}$ from a discretized time series $\mathcal{T}=\{ \langle ts_k,level_k \rangle \}_{k\in K\subset\nat}$.

$\mathcal{G}$ is generated considering {\em epochs}, i.e., consecutive time intervals of equal size $s$ (number of observations per epoch), where $s$ is an input parameter, see Algorithm~\ref{algo:teg_discovery}. 
Let $|K|=N$ be the total number of observations in $\mathcal{T}$, then the epochs are defined as follows:
\[ epoch_j = [ ts_{js}, ts_{(j+1)s}), \ \ j\in J = \{0,\dots, \lfloor N/s\rfloor -1\} \]

\begin{algorithm}[!htb]
\SetAlgoLined
\KwData{$\mathcal{T} = \{ \langle ts_k, level_k \rangle \}_{k\in K}$ (discretized time series), $s$ (num. of observations per epoch)}
\KwResult{$\mathcal{G} = \{ G_j \}_{j\in J}$ (time-evolving graph)}
$\mathcal{G} = \emptyset$\;
\ForEach{$j \in [0,\dots,\lfloor N/s\rfloor{-}1]$}{
	$\mathcal{T}_j$ = extractTimeSeries($\mathcal{T}$, $s$, $j$) \tcp*{time series in the interval $\big[js,(j+1)s\big)$} 
	$G_j$ = generateGraph($\mathcal{T}_j$)\;	
	append($G_j$, $\mathcal{G}$)\;
}
 \caption{Graphs discovery}
 \label{algo:teg_discovery}
\end{algorithm}

Therefore, a graph $G_j \in \mathcal{G}$ synthetizes the observations in an epoch $j\in J$.
Concretely, the nodes of $G_j$ represent the discretized observations in $j$, i.e., the levels, whereas the edges represent transitions (changes) between levels.  

Algorithm~\ref{algo:graph_generation} sketches the generation of a graph $G$, from the portion of the discretized time series $\mathcal{T}$ 
corresponding to an epoch of size $s$. 
The nodes $V = \{ \langle l_i, w_i \rangle\}$ are pairs, where $l_i$ is a level, taken from the observations in the considered epoch, and $w_i$ is the 
number of occurrences of $l_i$ in the epoch. $V$ is ordered according to the levels $l_i$, and its size $|V| \leq s$.
The edges $E$ define an adjacency matrix, where each entry $E_{rc}$ means the number of transition occurrences
from level $r$ to level $c$ in the epoch. Thus, each graph is directed and weighted.

\begin{algorithm}[!htb]
\SetAlgoLined
\KwData{$\mathcal{T} = \{ \langle ts_k, level_k \rangle \}_{k\in K'}$ (discretized time series with $|K'|=s$ observations)}
\KwResult{$G = (V,E)$ (weighted graph: weights are associated to both nodes and edges)}
levels = extractLevels($\mathcal{T}$)   \tcp*{levels is an ordered multiset of levels}
\tcp{$V$ is an ordered set of levels with their frequencies}
V = setNodes(levels, s)\; 
$E$ = initializeAdjacencyMatrix(|V|) \tcp*{$E$ is a $|V|\times |V|$ matrix with 0-entries}
\ForEach{$k \in [0,s{-}2]$}{
	r = getIndex(levels, k)\;
	c = getIndex(levels, k+1)\;
	$E_{r,c}$ = $E_{r,c}$ + 1\;
}
\caption{Graph generation}
\label{algo:graph_generation}
\end{algorithm}

\subsection{Model construction} 
A prediction model ${\mathcal M}(G,D_\mu)$ can be built using $\mathcal{G} = \{ G_j \}_{j\in J\subset\nat}$, a set of TEGs generated from a discretized training dataset. 
$G$ is a \emph{global graph} synthesizing the graphs in $\mathcal{G}$, and $D_\mu$ is a \emph{baseline distribution} of the dissimilarities between each graph $G_j$ and the global graph $G$. 
Since different dissimilarity metrics $\mu$ can be used to calculate this distribution, then there are just as many prediction models and, therefore, as many different anomaly detectors.

\subsubsection{Global graph}

A global graph is defined as the \emph{sum} of all the graphs in $\mathcal{G}$:
\begin{equation}\label{eq:globalGraph}
G  \stackrel{def}{=} \sum_{j=1}^{|J|} G_i = G_1 + G_2  + \dots + G_{|J|}
\end{equation}
where the \emph{sum} binary operator is defined as follows.

\paragraph{Sum of graphs}
Let $G_1(V_1,E_1)$ and $G_2(V_2,E_2)$ be two graphs, where $V_i = L_i\times \mathbb{N} (i=1,2)$ includes the set of pairs
$\langle l, w \rangle$, ordered according to the levels $l \in L_i$, and  $E_i (i=1,2)$ is the adjacency matrix representing the graph edges and their weights.

The sum of $G_1$ and $G_2$, denoted as $G_1 + G_2$, is a graph $G(V,E)$  where the set of nodes $V =(L_1\cup L_2)\times \mathbb{N}$ is:
\begin{eqnarray*}
 V  & = & \{ \langle l, w_1+w_2 \rangle | \langle l, w_1 \rangle \in V_1\ \land\   \langle l, w_2 \rangle \in V_2\} 
 			\cup V_1 \setminus V_2 \cup V_2\setminus V_1 
\end{eqnarray*}
The adjacency matrix $E = [e_{ij}]$ of $G$ is a matrix of dimension $|L_1\cup L_2|\times |L_1\cup L_2|$, with the following entries: 
\[ e_{ij} = \begin{cases}
 		x_{ij} + y_{ij} & \mathrm{if}\ i, j \in L_1 \cap L_2 \\
		x_{ij}		    & \mathrm{if}\ i,j \in L_1\ \land\ (i\not\in L_{2} \vee j \not\in L_{2})\\
		y_{ij}		    & \mathrm{if}\ i,j \in L_2\ \land\ (i\not\in L_{1} \vee j \not\in L_{1})
		\end{cases}
\]
where $E_1 = [ x_{ij}]$ and $E_2 = [y_{ij}]$.

\subsubsection{Baseline distribution}
Let $\mathcal{G}$ be a sequence of graphs, $G$ its global graph, as defined in Equation~\ref{eq:globalGraph},
and $\mu: \mathcal{G}\times\{ G \}\rightarrow \rea^+_0$ a dissimilarity metric.
Then, the baseline distribution is a function $D_\mu: J \rightarrow \rea^+_0$  defined as follows:
\begin{equation}\label{eq:baseline}
D_\mu(j) = \mu(G_j, G)
\end{equation}

There exists many metrics that can be used to compute dissimilarities between two graphs~\citep{ATK15,Cha07}. The {\bf tegdet} library currently implements $28$ different metrics, which are detailed in Appendix~\ref{app:technical}.

Prior to compute a dissimilarity between $G$ and a given $G_j$, it is mandatory to {\em resize} $G_j$.
This means to expand the set of nodes and the incidence matrix of $G_j$ to obtain a new graph with the same ordered set of nodes as in  $G$ (Algorithm~\ref{algo:graph_resizing}). 
In the expansion, the frequencies associated to the new matrix entries are set to a wildcard value, so to differentiate them from those with zero values, which indicate the absence of an edge but the presence of a node.
Such wildcard, will be treated differently, according to the concrete metric $\mu$.

\begin{algorithm}[!htb]
\SetAlgoLined
\KwData{$G_1 = (V_1,E_1)$, where $V_1=L_1\times W_1$ (graph to be resized)\;
$L$ (ordered set of levels of $G$)}
$wildcard = -1$\;
\ForEach{$level \in L$}{
	\If{$level\not\in L_1$}{ 
		\tcp{Expand the graph}
		 $position$ = getPosition($level$)\;
		 insertNewNode($V_1$, $position$, $\langle level, wildcard\rangle$)\;
		 insertNewColumn($E_1$, $position$, $wildcard$)\;
		 insertNewRow($E_1$, $position$, $wildcard$)\;
	}
}
\caption{Graph resizing}
\label{algo:graph_resizing}
\end{algorithm}

In the following, two different examples of baseline distributions, $D_\mu$, are presented. The first one uses the~\cite{Hamming} metric, which is based on the structure of the graphs. The second one uses the~\cite{Jeffreys61} metric, which is based on the edge relative frequencies.

\paragraph{Hamming dissimilarity distribution}
Formally, let $A = [a_{ij}]$ and $B = [b_{ij}]$ be the $n\times n$ adjacency matrices of the graphs to be compared,  
$G_A$ and $G_B$ respectively, where the wildcard entries are set to $-1$ value.

The Hamming dissimilarity distribution is defined as follows:
\begin{equation}\label{eq:Hamming}
\mu(G_A, G_B) =
1 - \frac{\mathop{\sum_{i=1}^n \sum_{j=1}^n} \chi(a_{ij},b_{ij})}{n^2}
\end{equation}
where  $\chi$ is the following indicator function:
\begin{equation*}
\chi(x,y) = 
\begin{cases}
1 & \mathrm{if\ } (x>0 \land y >0) \vee (x=0 \land y = 0)\\
0 & \mathrm{otherwise}
\end{cases}
\end{equation*}

\paragraph{Jeffreys dissimilarity distribution}
Formally, let $A = [a_{ij}]$ and $B = [b_{ij}]$ be the $n\times n$ adjacency matrices of the graphs to be compared,  
$G_A$ and $G_B$ respectively, where wildcard entries are set to zero value.
Let $A' = [a'_{ij}]$ and $B'=[b'_{ij}]$ be the matrices obtained from $A$ and $B$, respectively, 
where the entries represent relative frequencies:
\[  a'_{ij} = \frac{a_{ij}}{(\sum_{k=1}^n \sum_{l=1}^n a_{kl})}, \ 
    b'_{ij} = \frac{b_{ij}}{(\sum_{k=1}^n \sum_{l=1}^n b_{kl})}.
\]
The Jeffreys dissimilarity distribution is defined as follows:

\begin{eqnarray}\label{eq:KL}
\mu(G_A,  G_B) & = & \sum_{i=1}^{n} \sum_{j=1}^n (a'_{ij} - b'_{ij})\ log\ \frac{a'_{ij}}{b'_{ij}}
\end{eqnarray}

\subsection{Detection of outliers}

Prediction models ${\mathcal M}(G,D_\mu)$ are used to detect outliers in {\em testing} datasets.
A {\em testing} dataset $\mathcal{T'}$ is made of an ordered set of observations, accounting for the same phenomenon as the {\em training} dataset $\mathcal{T}$ used to build the model. Such observations should be taken with the very same time granularity as for the training dataset.

The anomaly detection phase implies the generation of TEGs from  $\mathcal{T'}$ as described in sub-Sections~\ref{SS:discretization} and~\ref{SS:discovery}.
Hence, the discretization of $\mathcal{T'}$ is carried out using the function~(\ref{e:discretization}), which is defined over the training dataset $\mathcal{T}$, whereas the graph discovery is performed considering the very same number of observations per epoch as in the model building phase. 

Therefore, let $\mathcal{G'} = \{ G'_j \}_{j\in J'\subset\nat}$ be a set of TEGs obtained from the testing dataset $\mathcal{T'}$, then: 
\begin{framed} 
The epoch $j\in J'$ is  \textbf{anomalous} if the dissimilarity between $G'_j$ and the global graph $G$, computed using
the metric $\mu$, is above  the $100{-}\alpha$ percentile of the baseline distribution $D_\mu$.
\end{framed}
Observe that  $\alpha\in(0,100)$ represents the significance level and it is an input parameter.
Algorithm~\ref{algo:outlierDetection} detects outlier epochs, given a prediction model, the significance level and 
the sequence of graphs generated from the training dataset.

\begin{algorithm}[!htb]
\SetAlgoLined
\KwData{${\mathcal M}(G,D_\mu)$ (prediction model), $\alpha$ (significance level),
 $\mathcal{G'} = \{ G'_j \}_{j\in J'\subset\nat}$ sequence of graphs from $\mathcal{T'}$}

$thresold$ = percentile($D_\mu, 100{-}\alpha$)\;
outlier = $\emptyset$\;
\ForEach{$j \in J'$}{
	\If{$\mu(G'_j,G)$ > thresold}{ 
		$outlier$ = $outlier \cup \{ j \}$\;
	}
}
\caption{Detection of outliers}
\label{algo:outlierDetection}
\end{algorithm}

\newpage
\section{Implementation and use of the library}\label{sec:impl}
\begin{figure}[hbt]
	\centering
	\includegraphics[width=0.6\textwidth]{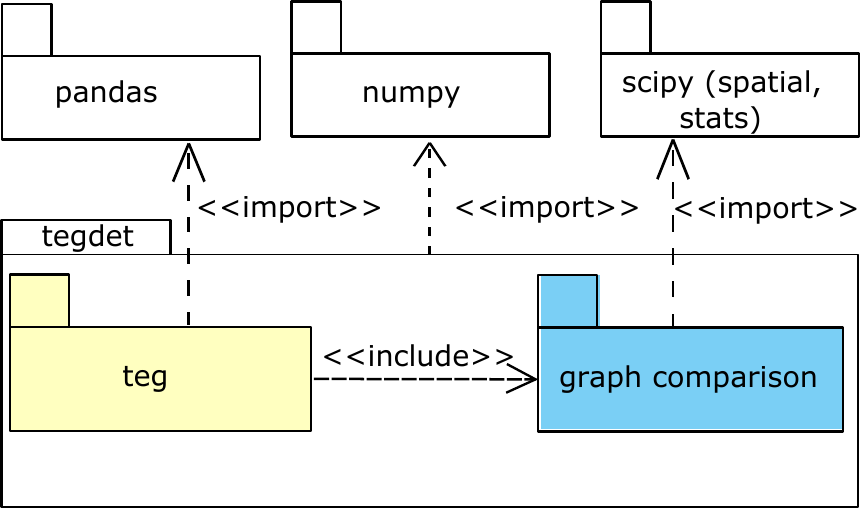}
	\caption{Overview of the {\bf tegdet} library}
	\label{fig:packageOverview}			
\end{figure}

The {\bf tegdet} library has been implemented in \proglang{Python} language~\citep{van2007python}. 
As shown in Figure~\ref{fig:packageOverview}, the library uses three other different \proglang{Python} libraries and it is made of two packages:
\begin{itemize}
    \item \emph{teg}: It is the main package. It defines the API for the users of the library.
    \item \emph{graph comparison}: It is responsible for creating graphs and computing the dissimilarity between two graphs, according to a given metric.
\end{itemize}

\begin{figure}[hbt]
	\includegraphics[width=\textwidth]{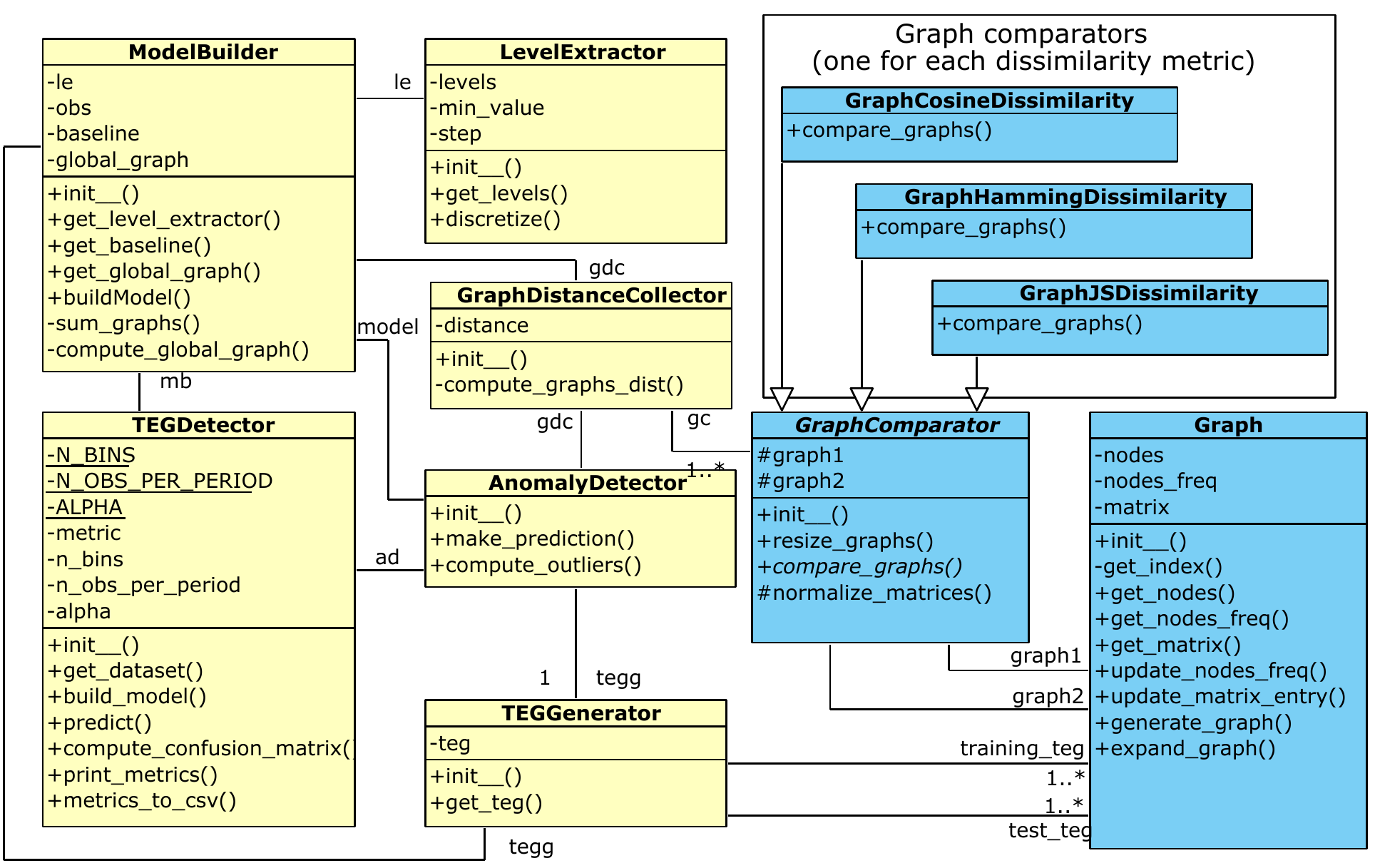}
	\caption{Class diagram of the {\bf tegdet}}
	\label{fig:tegdetCD}			
\end{figure}
 
Each package of the library is made up of a set of \proglang{Python} classes\footnote{A detailed description of the attributes and methods of these classes can be found in: \url{https://github.com/DiasporeUnizar/TEG}.}. Figure~\ref{fig:tegdetCD} depicts all the classes in the library, different colours are used to map classes to packages, according to Figure~\ref{fig:packageOverview}. In particular, the methods of the \texttt{TEGDetector} class conform the API of the \pkg{tegdet} library, they are described in Table~\ref{tab:TEGclassMethods}, Appendix~\ref{app:api}. 
The methods of the API are for the users of the library to construct the prediction model and carry out the anomaly detection. 
In the following, both scenarios are described.

\subsection{Construction of the prediction model}\label{subsec:construction}

\begin{figure}[hbt]
	\includegraphics[width=\textwidth]{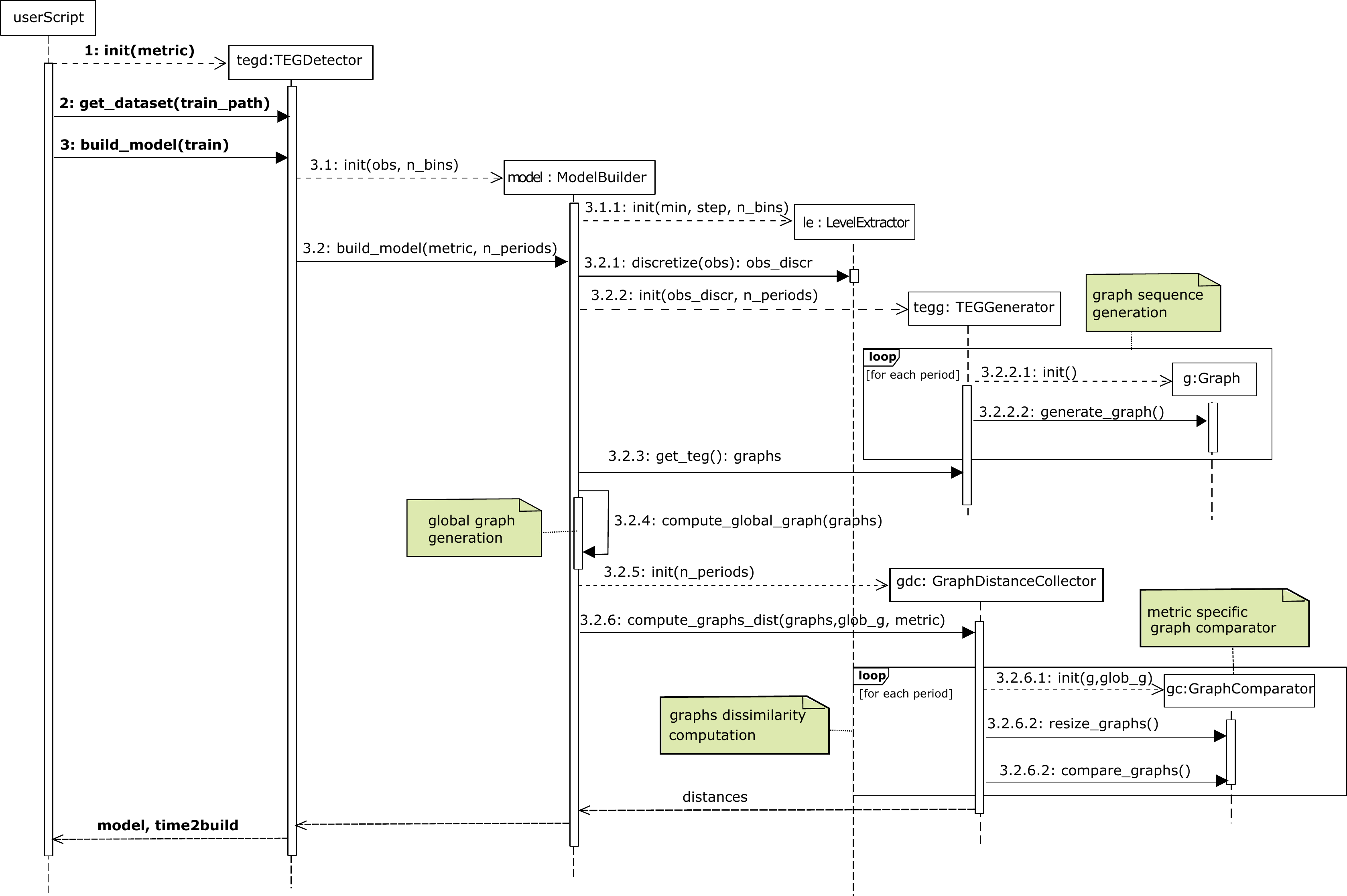}
	\caption{Model construction}
	\label{fig:modelConstruction}			
\end{figure}

Figure~\ref{fig:modelConstruction} shows the interaction of the \pkg{tegdet} classes to carry out the construction of a {\em prediction model}. 
The \emph{userScript} represents a \code{Python} script written by a user of the library. It must sequentially call three API methods. 
First, the \code{init} method sets some parameters needed to create the {\em training} TEGs and the prediction model. These parameters are detailed in Table~\ref{tab:TEGclassAttributes} (Appendix~\ref{app:api}). 
In Figure~\ref{fig:modelConstruction}, only the \code{metric} parameter is set since default values are left for the remaining parameters.
Next, the \code{get\_dataset} method loads the {\em training} dataset, as a \code{pandas DataFrame}. 
The third call, \code{build\_model}, actually generates the prediction model, as follows.

Initially, instances of the \code{ModelBuilder}  and \code{LevelExtractor} classes are created. 
Then, the \code{build\_model} call to the \code{ModelBuilder} instance: 
a) produces a discretized dataset (\code{discretize} call), according to Function~\ref{e:discretization}; 
b) generates a sequence of {\em training} graphs (generation of an instance of \code{TEGGenerator} class and
 first \code{loop} fragment), as described in Algorithm~\ref{algo:teg_discovery}; 
c) obtains the global graph (\code{compute\_global\_graph} call), according to Equation~\ref{eq:globalGraph}; and 
d) computes the dissimilarities between each graph of the sequence and the global graph (generation of an instance of \code{GraphDistanceCollector} and last \code{loop} fragment), following Equation~\ref{eq:baseline} and Algorithm~\ref{algo:graph_resizing}.
Finally, the results, that is the \code{model} and the time required to build it, are returned to the \emph{userScript}.

It is worth to observe that, although this scenario shows an instance of the class \code{GraphComparator}, 
within the last \code{loop} fragment, in general the computation of graphs dissimilarities is carried out by a sub-class of \code{GraphComparator} (cfr. Figure~\ref{fig:tegdetCD}, sub-classes of \emph{GraphComparator}), in fact the sub-class corresponding to the metric to be computed.

\subsection{Anomaly detection}\label{subsec:detection}

\begin{figure}[hbt]
	\includegraphics[width=\textwidth]{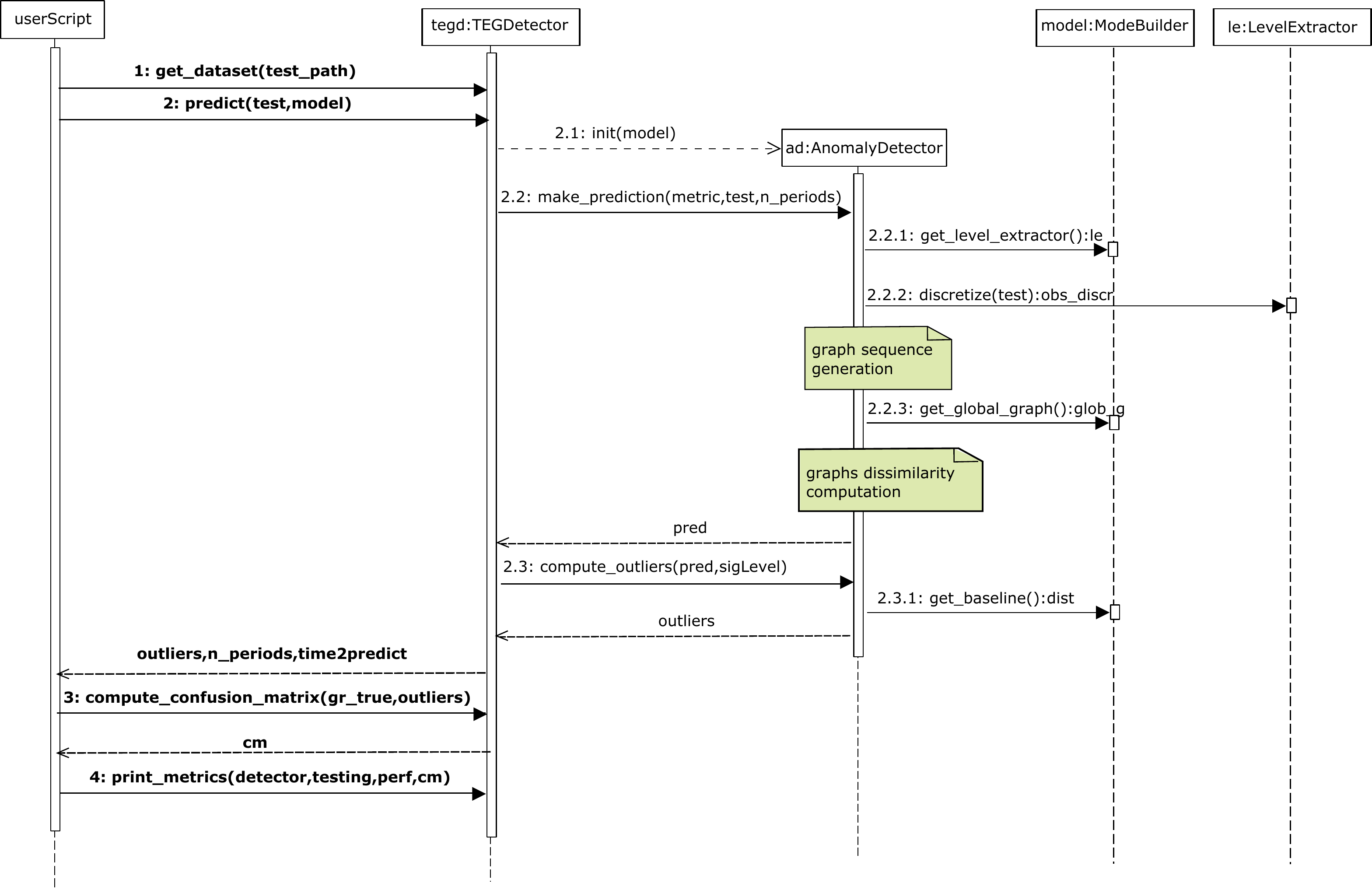}
	\caption{Anomaly detection and metric printing}
	\label{fig:anomalyDetection}			
\end{figure}

Figure~\ref{fig:anomalyDetection} shows the interaction of the \pkg{tegdet} classes to carry out the anomaly detection. First, observe that instances of \code{TEGDetector}, \code{ModelBuilder} and \code{LevelExtractor} have been already created in the model phase. The \emph{userScript} starts the interaction by calling \code{get\_dataset} to load the {\em testing} dataset as a \code{pandas DataFrame}.  
Then, the \code{predict} call triggers three sub-steps: the creation of an instance of \code{AnomalyDetector},  \code{make_prediction} and \code{compute_outliers}, which evolve as follows.

Predictions are carried out by: 
a) retrieving the discretization levels from the \code{ModelBuilder} and creating the discretized {\em testing} dataset (\code{discretize} call), 
according to Function~\ref{e:discretization}; 
b) generating the corresponding  sequence of {\em testing} graphs, as described in Algorithm~\ref{algo:teg_discovery};  and 
c) computing the graph dissimilarities, following Equation~\ref{eq:baseline} and Algorithm~\ref{algo:graph_resizing}. 
Dissimilarities are computed using the global graph, retrieved from the \code{ModelBuilder}, and the metric selected in the first step. 
Then, the computation of outliers (Algorithm~\ref{algo:outlierDetection}), i.e., the detection of anomalous epochs, is carried out considering the predictions and the baseline distribution, which is also retrieved from the \code{ModelBuilder}. 

Next, the scenario computes a confusion matrix (\code{compute\_confusion\_matrix} call), that can be used to 
assess the prediction capabilities of the anomaly detector. In particular,  its entries summarize the results of the anomaly detection as follows:
\begin{itemize}
	\item True positives {\em tp}, i.e., number of epochs that are \textbf{correctly} predicted as \textbf{anomalous}; 
	\item True negatives {\em tn}, i.e, number of epochs that are \textbf{correctly} predicted as \textbf{not anomalous};
	\item False positives {\em fp}, i.e., number of epochs that are \textbf{incorrectly} predicted as \textbf{anomalous}; and
	\item	False negatives {\em fn}, i.e., number of epochs that are \textbf{incorrectly} predicted as \textbf{not anomalous}.
\end{itemize}
The scenario ends by printing, in the standard output, the confusion matrix, the time needed to build the model and to make the prediction  
(\code{print\_metrics} call). 

\newpage
\section{Experiment: Smart grid fraud detection} \label{sec:application}

Smart grids are complex cyber-physical systems built on top of electrical power infrastructures. Although resilient and reliable to supply electricity, smart grids are vulnerable to attacks and frauds~\citep{IET16}.~\cite{BJMR21} studied different types of attacks on smart grids and evaluated the effectiveness of some anomaly detection techniques for these attacks. Now, we borrow some experiments in~\cite{BJMR21} to illustrate how to build a prediction model and detect anomalies using the \pkg{tegdet} library.

\subsection{Description of the dataset}\label{ss:dataset}

The work of~\cite{BJMR21} used datasets from the Ireland’s Commission for Energy Regulation~\citep{ISSDA-CER}.
From them, synthetic datasets were created to model different types of attacks. 
Here, we use one of these synthetic datasets and some information related to a specific smart meter~\citep{ISSDA-CER}.

Concretely, our dataset is made of real value observations, representing a smart meter energy consumption, in kWh, collected every half hour.
The observation period is $75$ weeks.
The dataset is organized in three files\footnote{The dataset is available in the GitHub repository:
\url{https://github.com/DiasporeUnizar/TEG}}:
\begin{itemize}
 \item \code{training}: Includes readings for the first $60$ weeks. It will be used to build prediction models.
 \item \code{test\_normal}: Includes readings for the remaining $15$ weeks. Its observations are considered \emph{normal}, i.e., it is assumed that they have not been affected by attacks. It will be used to detect anomalies. 
\item \code{test\_anomalous}: This is a synthetic dataset created from the \code{test\_normal} file, so it includes readings for the same period.
It is assumed that its observations have been tampered to defraud the energy utility by paying less than consumed. It will also be used to detect anomalies.
\end{itemize}

\begin{figure}[htb]
\begin{minipage}{0.65\textwidth}
	\includegraphics[width=\textwidth]{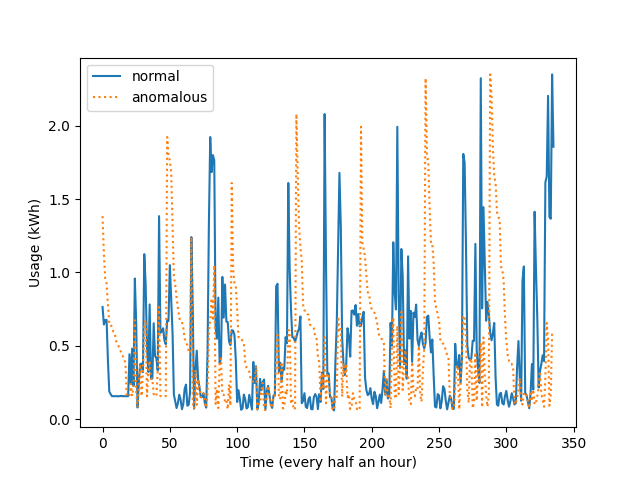}
\end{minipage}\
\begin{minipage}{0.3\textwidth}
\begin{framed}
\begin{verbatim}
DT,Usage
19501,0.646
19502,0.494
19503,0.623
19504,0.618
19505,0.587
19506,0.317
19507,0.13
19508,0.056
19509,0.133
...
\end{verbatim}
\end{framed}
\end{minipage}
\caption{(left) Time series of the two testing sets, one week observations. (right) An excerpt of the training dataset}
\label{fig:dataset}	
\end{figure}

Figure~\ref{fig:dataset} (left) represents the observations in the first week of \code{test\_normal} and \code{test\_anomalous}. The \emph{anomalous}
curve has been obtained by swapping the observations between peak and peak-off periods (from 9:00am to midnight and from midnight to 9:00am), so assuming to defraud a time-of-use contract having peak and off-peak periods~\citep{KLWIS16}.
Figure~\ref{fig:dataset} (right) offers an excerpt of the \code{training} file, which is in \texttt{csv}\footnote{Comma-separated value.} format. The first column indicates the time instant (observations are time ordered) and the second one the observed real value at that instant. 
The first row indicates the column headers. The \pkg{tegdet} library does not impose any restrictions on column names or time instant format.

\subsection{How to build a prediction model and detect anomalies}\label{ss:variantsAnalysis}
The scenarios presented in Subsections~\ref{subsec:construction}~and~\ref{subsec:detection} are now developed using \proglang{Python} code to illustrate how to carry out an experiment with the \pkg{tegdet} library. 
In particular, the \proglang{Python} script in Appendix~\ref{app:scripts}, Listing~\ref{list:tegdet_variants}, is used. 
Moreover, Appendix~\ref{app:install} offers instructions for the installation of the required libraries to execute the script. 
 
Initially, the script imports the API class \code{TEGDetector} from the \pkg{tegdet} library:
\begin{lstlisting}[numbers=none,backgroundcolor = \color{vlgray}]
from tegdet.teg import TEGDetector
\end{lstlisting}

Then, the script sets the paths for the datasets and a results file:
\begin{lstlisting}[numbers=none,backgroundcolor = \color{vlgray}]
TRAINING_DS_PATH = "/dataset/energy/training.csv "
TEST_DS_PATH = "/dataset/energy/test_"
RESULTS_PATH = "/script_results/energy/tegdet_variants_results.csv"
\end{lstlisting}

The script defines also the testing datasets and metrics that will be used: 
\begin{lstlisting}[numbers=none,backgroundcolor = \color{vlgray}]
list_of_testing = ("normal", "anomalous")
list_of_metrics = ("Hamming", "Cosine", "Jaccard", "Dice", "KL", "Jeffreys", "JS", 
	"Euclidean", "Cityblock", "Chebyshev", "Minkowski", "Braycurtis", "Gower",
	"Soergel", "Kulczynski", "Canberra", "Lorentzian", "Bhattacharyya", 
	"Hellinger", "Matusita", "Squaredchord", "Pearson", "Neyman", "Squared",
	"Probsymmetric", "Divergence","Clark", "Additivesymmetric" )
\end{lstlisting}

The preliminaries of the script end by setting the input parameters, as indicated in Table~\ref{tab:TEGclassAttributes} (Appendix~\ref{app:api}). 
Concretely, the energy consumption range is partitioned into $30$ discretization levels, the epochs correspond to weeks ($336$ observations per period), and the predictions will be made considering a $95\%$ significance level:
\begin{lstlisting}[numbers=none,backgroundcolor = \color{vlgray}]
n_bins = 30
n_obs_per_period = 336
alpha = 5
\end{lstlisting}

\paragraph{Building the prediction model.}
The script builds a prediction model for a given metric in four steps: 
1) it stores the path to the training dataset, 2) instantiates the \texttt{TEGDetector} class with the current metric, 3) loads the training dataset, and 4) builds the model: 
\begin{lstlisting}[numbers=none,backgroundcolor = \color{vlgray}]
def build_and_predict(metric):
	cwd = os.getcwd() 
	train_path = cwd + TRAINING_DS_PATH
	tegd = TEGDetector(metric)
	train = tegd.get_dataset(train_path)
	model, time2build = tegd.build_model(train)
\end{lstlisting}

The output of the training dataset shows that the column headers have been renamed (\code{TS}, time-stamp and \code{DP}, data-point),
and that it includes $20160$ observations, during an overall period of $60$ weeks:
\begin{framed}
{\small
\begin{CodeOutput}
>>> train
          TS     DP
0      19501  0.646
1      19502  0.494
2      19503  0.623
3      19504  0.618
4      19505  0.587
...      ...    ...
20155  62144  0.550
20156  62145  0.586
20157  62146  0.585
20158  62147  1.898
20159  62148  0.988
[20160 rows x 2 columns]
\end{CodeOutput}
}
\end{framed}

Also as a result, the time to build the model (in seconds) is stored:
\begin{framed}
{\small
\begin{CodeOutput}
>>> time2build
0.1602308750152588
\end{CodeOutput}
}
\end{framed}

\paragraph{Anomaly detection.}

For each testing dataset, the script: 1) stores its path, 2) loads the dataset through the \texttt{TEGDetector} instance, 
and 3) carries out prediction for the current dataset and model:
\begin{lstlisting}[numbers=none,backgroundcolor = \color{vlgray}]
	for testing in list_of_testing:
		test_path = cwd + TEST_DS_PATH + testing + ".csv" 
		test = tegd.get_dataset(test_path)
		outliers, n_periods, time2predict = tegd.predict(test, model) 
\end{lstlisting}
As a result of the prediction, the script obtains: a) the \code{outliers}, an array where each entry corresponds to the prediction related to an epoch
(i.e., $0$ normal epoch, $1$ anomalous epoch), b) the number of  epochs (\code{n\_periods}) and c) the time required to make the prediction (\code{time2predict}, in seconds).

The following snapshot presents the prediction of the Hamming detector when applied to the \code{normal} dataset. 
It detects one anomalous epoch in the overall set of $15$  weeks, concretely the $4^{th}$ week (i.e., the $4^{th}$ entry in the array):
\begin{framed}
{\small
\begin{CodeOutput}
>>> outliers
array([0., 0., 0., 1., 0., 0., 0., 0., 0., 0., 0., 0., 0., 0., 0.])
\end{CodeOutput}
}
\end{framed}
Next, the script collects results of the predictions. First, it computes the confusion matrix, using ground true values for each epoch, i.e., weeks. These values are obtained using the \code{ones} or \code{zeros} functions of the \pkg{numpy} library. Concretely,  the \code{normal} dataset ground values are set to zero (i.e., its observations correspond to actual energy consumption), whereas the \code{anomalous} dataset ground values are set to one (i.e., its observations correspond to tampered consumptions):
\begin{lstlisting}[numbers=none,backgroundcolor = \color{vlgray}]
		if testing == "anomalous":
			ground_true = np.ones(n_periods)        
		else:
			ground_true = np.zeros(n_periods)            
		cm = tegd.compute_confusion_matrix(ground_true, outliers)
\end{lstlisting}
The following snapshot presents the confusion matrix \code{cm} of the Hamming detector when applied to the \code{normal} testing dataset. 
A week is wrongly detected as anomalous (\code{fp} entry), while the rest of the weeks are correctly predicted as normal weeks (\code{tn} entry):
\begin{framed}
{\small
\begin{CodeOutput}
>>> cm
{'tp': 0, 'tn': 14, 'fp': 1, 'fn': 0}
\end{CodeOutput}
}
\end{framed}
Finally, the script: 1) prints, in the standard output, the confusion matrix and performance metrics, i.e., the time to build the model and the time to make prediction, 
and 2) stores the results in a \pkg{csv} file for post-processing purposes:
\begin{lstlisting}[numbers=none,backgroundcolor = \color{vlgray}]
 		perf = {'tmc': time2build, 'tmp': time2predict}
		tegd.print_metrics(detector, testing, perf, cm)
		results_path = cwd + RESULTS_PATH
		tegd.metrics_to_csv(detector, testing, perf, cm, results_path)
\end{lstlisting}
The following snapshot presents the results of the Hamming detector, considering both testing sets:
\begin{framed}
{\small
\begin{CodeOutput}
Detector:			    Hamming
N_bins:			      30
N_obs_per_period:		    336
Alpha:			       5
Testing set:			  normal
Time to build the model:	      0.1602308750152588 seconds
Time to make prediction:	      0.04300808906555176 seconds
Confusion matrix:	
 {'tp': 0, 'tn': 14, 'fp': 1, 'fn': 0}
Detector:			     Hamming
N_bins:			       30
N_obs_per_period:	             336
Alpha:			        5
Testing set:			  anomalous
Time to build the model:	      0.1602308750152588 seconds
Time to make prediction:	      0.059934139251708984 seconds
Confusion matrix:	
 {'tp': 10, 'tn': 0, 'fp': 0, 'fn': 5}
 \end{CodeOutput}
}
\end{framed}
The following snapshot provides an excerpt from the results file when applying the Hamming detector to both testing datasets:
\begin{framed}
{\footnotesize
\begin{CodeOutput}
detector,n_bins,n_obs_per_period,alpha,testing_set,time2build,time2predict,tp,tn,fp,fn
Hamming,30,336,5,normal,0.1602308750152588,0.04300808906555176,0,14,1,0
Hamming,30,336,5,anomalous,0.1602308750152588,0.059934139251708984,10,0,0,5
\end{CodeOutput}
}
\end{framed}


\section{Assessment of the library}\label{ss:basic_assessment}

The results file produced by the \pkg{tegdet} library can be post-processed to assess the quality of the very same library.
In the following, using the experiment in previous section we assess: a) the performance of the algorithms implemented to build the prediction model 
and to predict the anomaly; and b) the {\em accuracy} of the anomaly detectors.

Concerning performance, Table~\ref{tab:TegVariantResults} offers results regarding execution times\footnote{The analysis has been carried out using a laptop with Intel Core i7 double core CPU, 16GB RAM, 250GB SSD.}. 
On the one hand, we see that the mean time to build a model is approximately four times greater than the mean time to make a prediction. In fact, this is also the ratio  between the lengths of the training dataset ($60$ weeks) and each testing dataset ($15$ weeks).
On the other hand, the small standard deviations indicate that the times are not affected by the type of metric used
to build the anomaly detector.

\begin{table}[htb]
\begin{center}
\begin{tabular}{ccc}
\toprule
\textbf{Statistical qualifier} & \textbf{Time to build (ms.)} & \textbf{Time to predict (ms.)}\\\midrule
mean	& 174.46 & 47.70\\
std		&   11.17 &  3.58\\
min		& 160.23 & 44.31\\
max		& 215.39 & 62.16\\\bottomrule
\end{tabular}
\end{center}
\caption{Execution times statistics}
\label{tab:TegVariantResults}
\end{table}

\begin{figure}[htb]
\begin{center}
	\includegraphics[width=\textwidth]{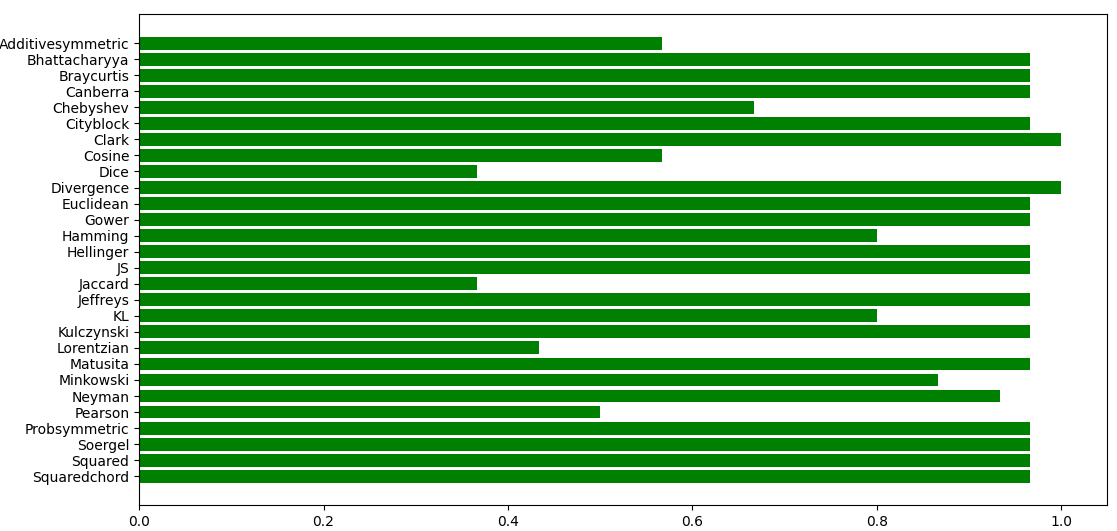}
\end{center}
\caption{Accuracy of the anomaly detectors}
\label{fig:Figure_2}	
\end{figure}

Figure~\ref{fig:Figure_2} shows the {\em accuracy} of the different anomaly detectors, which is defined as the ratio of correct predictions to total predictions:
\begin{equation}\label{eq:accuracy}
accuracy = \frac{(tp+tn)}{(tp+tn+fp+fn)}
\end{equation}
For each anomaly detector, the accuracy has been computed considering the sum of two confusion matrices, obtained from the 
\emph{normal} and \emph{anomalous} testing sets, respectively. Observe that the resulting confusion matrix is balanced (i.e, $tp+fp=fn+tn$).
Most of the detectors have high accuracy ($\geq 80\%)$, being Clark and Divergence the bests ($100\%$ of correct predictions).
Whereas, Dice ($37\%$), Jaccard ($37\%$) and Lorentzian ($43\%$) exhibit the least accurate predictions for the considered datasets.

\section{Guidelines for improving predictions}\label{sec:guidelines_predictions}
 
 An anomaly detector in \pkg{tegdeg} is characterized, besides the metric, by three  parameters:
1) number of observations per period, 2) number of discretization levels and 
3) $\alpha$, significance level in the detection phase.
An adequate setting of these parameters greatly impacts on the accuracy of the anomaly prediction, as well as on the time to build the model and to predict the anomaly.
This section provides guidelines for such adequate setting of the parameters.  

Table~\ref{tab:paramsRanges} shows the ranges considered from now on. The configuration used in Section~\ref{ss:basic_assessment} corresponds to the values in bold.
\begin{table}[htb]
\begin{center}
\begin{tabular}{ccc}
\toprule
\textbf{Parameter} & \textbf{Range} \\\midrule
\code{n_obs_per_period}	& [24, 48, 96, 168, 192, \textbf{336}, 480, 672, 816, 1008]\\
\code{n_bins}			& [5, 10, 15, 20, 25, \textbf{30}, 35, 40, 45, 50]\\
\code{alpha}			& [1, 2, 3, 4, \textbf{5}, 6, 7, 8, 9, 10]\\\bottomrule
\end{tabular}
\end{center}
\caption{Parameters ranges}
\label{tab:paramsRanges}
\end{table}

\paragraph{Performance.}

The times to build the model and predict the anomaly are mainly affected by the values of the \code{n_obs_per_period} and \code{n_bins} parameters. They
define the characteristics of the graph sequences and the graph dissimilarity distribution.

\begin{figure}[!htb]
     \centering
     \begin{subfigure}[b]{0.49\textwidth}
         \centering
         \includegraphics[width=\textwidth]{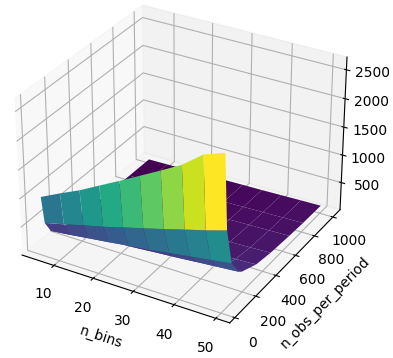}
         \caption{Mean time to build the model (ms.)}
         \label{fig:Figure_8_Hamming}        
     \end{subfigure}
     \hfill
     \begin{subfigure}[b]{0.47\textwidth}
         \centering
         \includegraphics[width=\textwidth]{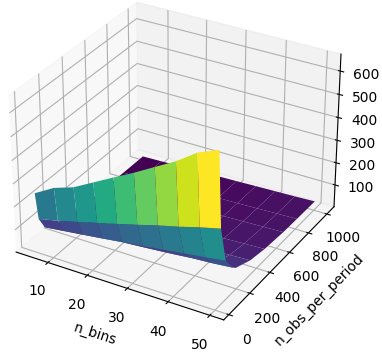}
         \caption{Mean time to make prediction (ms.)}
         \label{fig:Figure_9_Hamming}
     \end{subfigure}
 \caption{Performance results of the Hamming detector}
 \label{fig:times-Hamming}
\end{figure}

Figure~\ref{fig:times-Hamming} offers 3D plots of the mean execution times of the Hamming detector considering these two parameters. 
The two surfaces have a similar shape, despite the time ranges, which confirm the results in previous section, i.e., 
the times to build are approximately four time greater than the times to predict.
The rest of the detectors exhibit similar results.

We also observe that execution times are sensitive to both the parameters.
In particular, the lower is the value of  \code{n_obs_per_period} the higher is the execution time, since the longer will be the sequence 
of graphs to be generated.
For example, changing the value of  \code{n_obs_per_period} from $336$ to $24$ 
corresponds to change the epoch from $1$ week to $12$ hours, then the number of graphs to be generated for the training set 
will be $840$, instead of $60$ (for each testing set, $210$ graphs instead of $15$).
Besides, the higher is the value of  \code{n_bins} the higher is the execution time, since the size of the graphs to be generated (number of nodes and
edges), as well as the computation of the graph dissimilarity, are in direct proportion to the number of discretization levels.

\paragraph{Prediction accuracy.}

The accuracy is computed as in Equation~\ref{eq:accuracy} applied to the \emph{normal} and \emph{anomalous} testing datasets.

\begin{figure}[!htb]
\centering
      \begin{subfigure}[b]{0.41\textwidth}
         \centering
         \includegraphics[width=\textwidth]{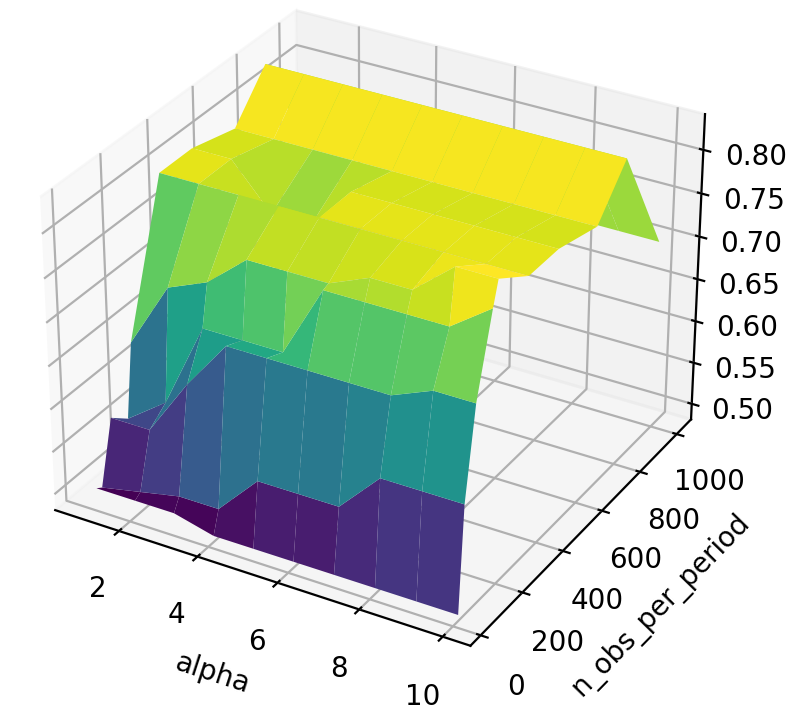}
         \caption{\texttt{n\_bin=30}}
         \label{fig:Figure_10_Hamming}
     \end{subfigure}
\hfill
     \begin{subfigure}[b]{0.40\textwidth}
         \centering
         \includegraphics[width=\textwidth]{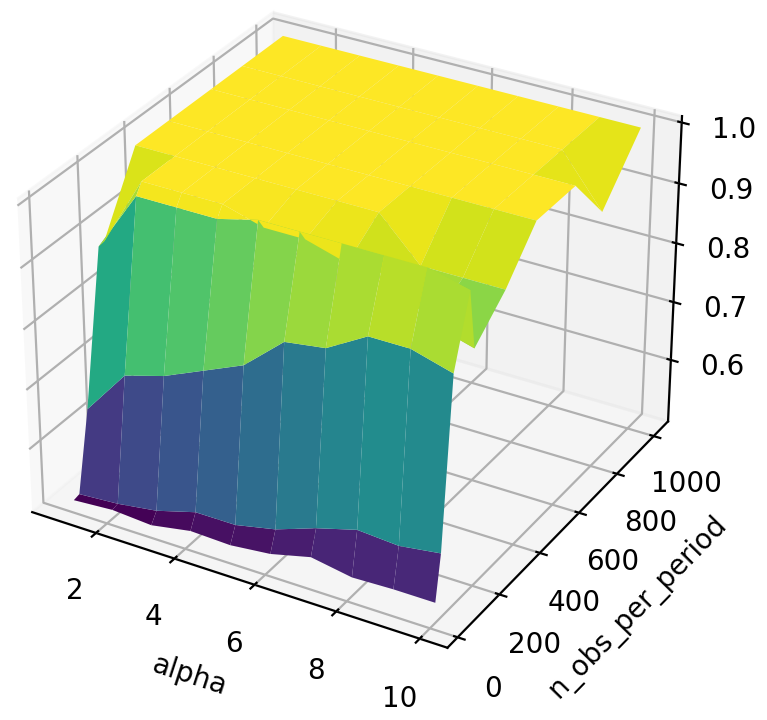}
         \caption{\texttt{n\_bin=30}}
         \label{fig:Figure_15_Clark}
     \end{subfigure}
\hfill
     \begin{subfigure}[b]{0.41\textwidth}
         \centering
         \includegraphics[width=\textwidth]{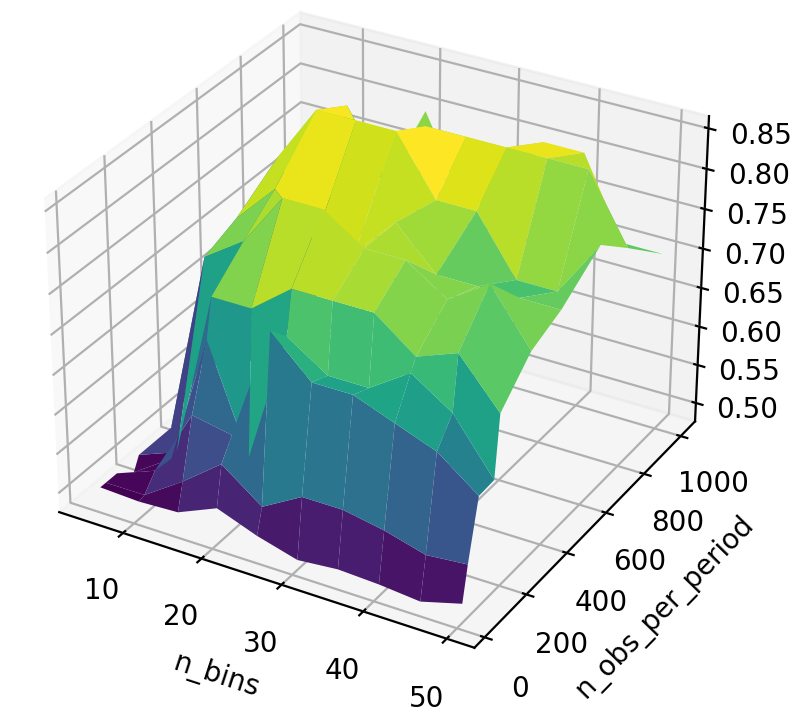}
         \caption{\texttt{alpha=5}}
         \label{fig:Figure_11_Hamming}
     \end{subfigure}
\hfill
     \begin{subfigure}[b]{0.40\textwidth}
         \centering
         \includegraphics[width=\textwidth]{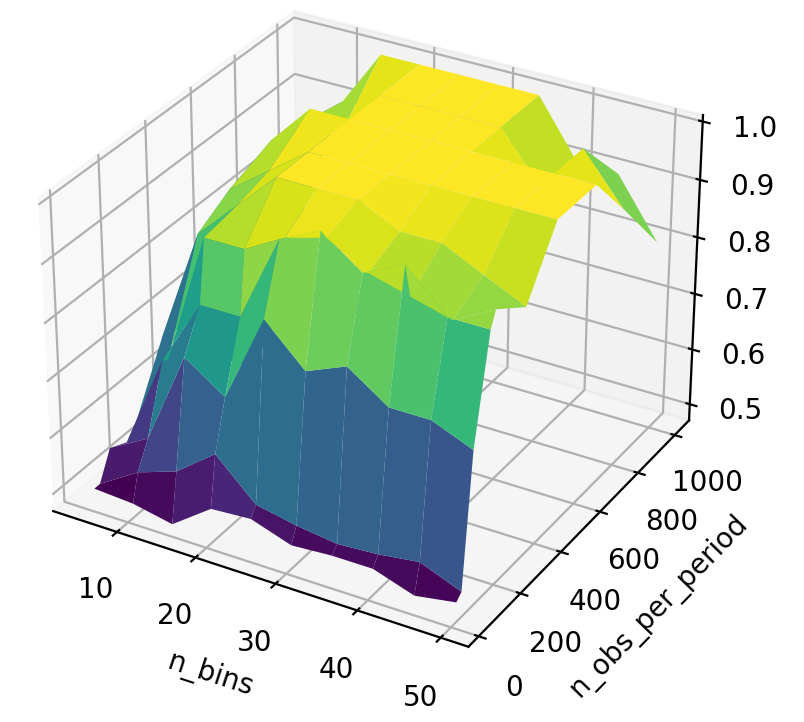}
         \caption{\texttt{alpha=5}}
         \label{fig:Figure_16_Clark}
     \end{subfigure}
\hfill
     \begin{subfigure}[b]{0.41\textwidth}
         \centering
         \includegraphics[width=\textwidth]{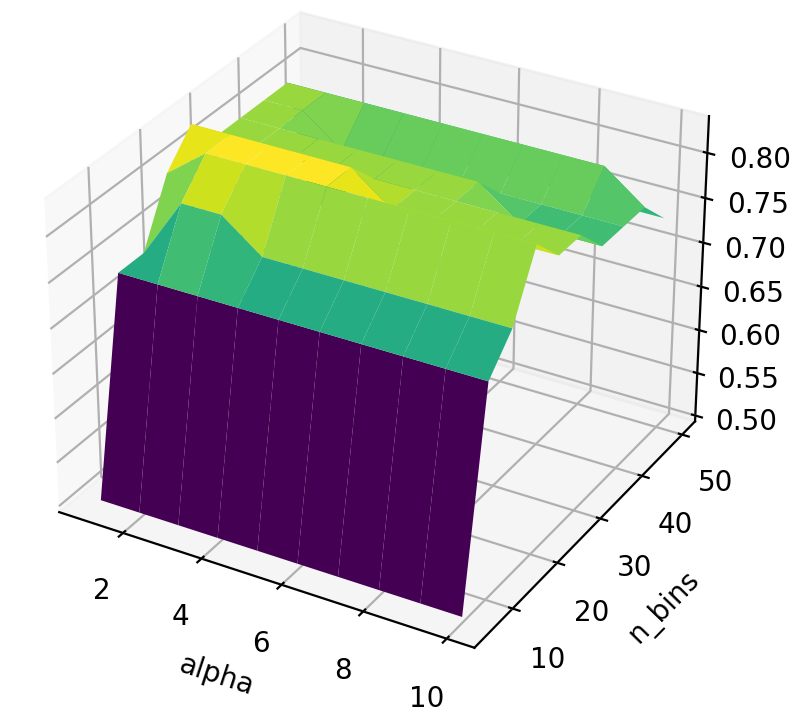}
         \caption{\texttt{n\_obs\_per\_period=336}}
         \label{fig:Figure_12_Hamming}
     \end{subfigure}
\hfill
     \begin{subfigure}[b]{0.40\textwidth}
         \centering
         \includegraphics[width=\textwidth]{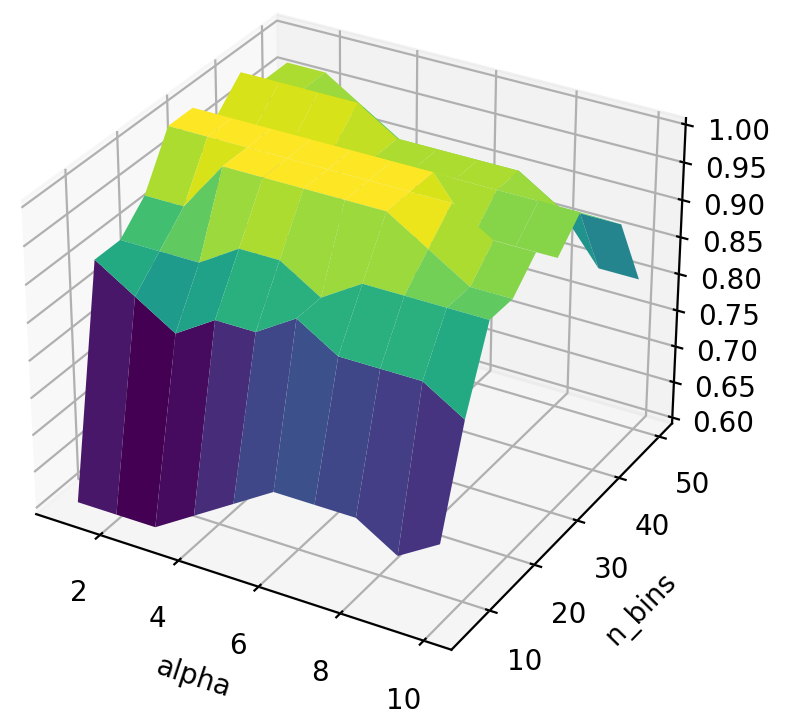}
         \caption{\texttt{n\_obs\_per\_period=336}}
         \label{fig:Figure_17_Clark}
     \end{subfigure}
\caption{Accuracy results of detectors: (left side) Hamming  and (right side) Clark}
\label{fig:accuracy}
\end{figure}
Figure~\ref{fig:accuracy} shows 3D plots for the accuracy of two anomaly detectors, 
Hamming (plots on the left) and Clark (plots on the right).
Each row considers the accuracy versus two of the three parameters. Then,
fixing the third parameter to its default value (bold value in Table~\ref{tab:paramsRanges}). 

Overall, the experiments confirm the results in the Figure~\ref{fig:Figure_2} for the default configuration, i.e., Clark outperforms Hamming. 
Indeed, the former is characterized by an accuracy range higher than the latter (cfr. the z-axis of the left plots vs/ the right ones).
For some configurations, Clark accuracy reaches 100\%, whereas the best value reached by Hamming is 85\% (Figure~\ref{fig:Figure_11_Hamming}).
Moreover, Clark seems to be more stable than Hamming, since its surfaces are more flattened than the Hamming ones.

Let us consider the first row. Figures~\ref{fig:Figure_10_Hamming} and~\ref{fig:Figure_15_Clark} 
show the accuracy versus the \code{alpha} and \code{n_obs_per_period} parameters. 
For both detectors, the accuracy seems to be slightly sensitive to the \code{alpha} parameter, which is used
to compute the threshold for the anomaly detection phase), whereas it is highly sensitive to the \code{n_obs_per_period}.
In particular, the lengths of the epochs between half-week ($168$ observations) and two weeks ($672$ observations) 
ensure better accuracy for both variants.

The second row shows the accuracy versus  \code{n_bin} and \code{n_obs_per_period} parameters 
(Figures~\ref{fig:Figure_11_Hamming} and~\ref{fig:Figure_16_Clark}).
Both surfaces are characterized by a higher slope in the direction of \code{n_obs_per_period}, meaning that the influence
of the second parameter is predominant regarding the first one.
The accuracy is also sensitive to the  \code{n_bin} parameter. In particular, values in the ranges $[15,30]$  and $[20,35]$ 
provide better accuracy for Hamming and Clark, respectively.

Finally, the third row shows the accuracy versus \code{alpha} and \code{n_bin} parameters.
Concerning Hamming, Figure~\ref{fig:Figure_12_Hamming} confirms that the variability with respect to the \code{alpha} values
is very low. Whereas, in the case of Clark (Figure~\ref{fig:Figure_17_Clark}), the accuracy is slightly sensitive to this parameter and 
values between $[3,6]$ provide better results.


\section{Conclusion} \label{sec:summary}

This first version of the \pkg{tegdet} library makes publicly available a \proglang{Python} API that enables to detect anomalies in univariate time series, by leveraging TEGs. 

The \pkg{tegdet} library decouples the creation of the TEGs and the anomaly detection. This is important for two reasons. First, this is what promotes the extensibility of the library, allowing to implement new anomaly detection techniques, while the discretization and graph discovery steps remain the same. Appendix~\ref{app:extensibility} explains how to accomplish the extensibility. Second, the decoupling also favours the good performance results of the library. Being the temporal complexity to create the TEGs linear, with respect to the length of the input, in the worst case, then, the performance impact of adding a new detection technique is attributable to the implementation of the technique exclusively.  

However, this version of the library has also some limitations. 
The library currently detects anomalous epochs, i.e., graph-level anomalies. 
Next extensions will identify which parts of the anomalous graph contribute the most to the change, i.e., attribution.  
The input of the library is currently a univariate time series, hence multivariate time-series should be further investigated.

Going back to the library performance, the most time consuming activities belong to the anomaly detection:  model construction and outlier detection. The worst case temporal complexity is $O(m |E|)$, where $m$ is the number of graphs and $|E|$ the size of the global graph (number of edges). 
$m$ depends on the length of the time-series $N$ and the number of observations per epoch $s$ (i.e., $m = \lfloor\frac{N}{s}\rfloor$), whereas $|E|$ depends on the
number of discretization levels $L$ and can be bounded by $L{-}1 \leq |E|\leq L^2$.
Therefore, the implemented anomaly detection technique is linear only with respect to the graph size 
(with a fixed $m$), as also proved by the results of the experiments --- cfr. in Figure~\ref{fig:times-Hamming}, 
the slope of the curves with respect to the axis \texttt{n\_bins} (i.e., $L$) and \texttt{n\_obs\_per\_period} (i.e., $s$).
We are investigating whether graphs sparse matrix implementations could even improve the (time and memory) efficiency of graph operations, such as the computation of graph dissimilarities.

The experiments presented here aim to provide guidelines for setting the parameters of the library.
Therefore, the purpose is not to make a robust evaluation of the quality of the detectors for a specific application domain.
Indeed, we have obtained the prediction accuracy results by considering a single time-series, i.e., the consumptions registered by one smart meter and a specific type of attack. Hence, although some detectors exhibit very good results, we cannot draw a general conclusion about their effectiveness. Furthermore, it is well-known that each application domain favours certain classes of detectors.

Finally, it is worth to say that our work is inspired by~\cite{MWXYZSXA21}, they propose a unified framework aimed at detecting different types of anomalies, for a specific application domain.
In our view, the framework shall include datasets, synthetic dataset generators and implementations of detectors, and shall provide support for the comparative analysis of the quality of the detectors. The \pkg{tegdet} library could be part of such framework.

\section*{Acknowledgments}
S. Bernardi and J. Merseguer were supported by the Spanish Ministry of Science and Innovation [ref. PID2020-113969RB-I00].




\newpage

\begin{appendix}


\section{Dissimilarity metrics} \label{app:technical}

Table~\ref{tab:dissimilarityMetrics} lists the metrics currently implemented in the \pkg{tegdet} library, which are used to compute dissimilarities between two graphs. 
The table groups the metrics by families, as given by~\cite{Cha07}. All the metrics are defined based on two $n$-length vectors, say $P$ and $Q$.
When applied to graph dissimilarity computation, the \emph{Hamming} metric considers only the structural characteristics of the graphs,  
then $P$ and $Q$ are obtained by just flattening the adjacency matrix of each graph.
In the case of the \emph{Cosine} metric,  $P$ and $Q$ are obtained by flattening the node frequency vector and the adjacency matrix of each graph.
Finally, for the rest of the metrics,  $P$ and $Q$ are obtained by flattening the adjacency matrix of each graph, 
where the entries represent absolute frequencies, and converting the resulting vectors into relative frequencies. 

\begin{longtable}{|l|c|}
    \toprule
    \textbf{Metric}  & \textbf{Definition}       \\ \midrule
    \multicolumn{2}{|c|}{Graph edit distance family}\\ \midrule
    Hamming & $1- \frac{\sum_{i=1}^{n} \chi(P_i, Q_i)}{n}$ \\ 
    			& where $\chi(x,y) = 
					\begin{cases}
						1 & \mathrm{if\ } (x>0 \land y >0) \vee (x=0 \land y = 0)\\
						0 & \mathrm{otherwise}
					\end{cases}
			$\\ \midrule
    \multicolumn{2}{|c|}{Inner product family} \\ \midrule
    Cosine  & $1- \frac{\sum_{i=1}^{n} P_i\cdot Q_i}{\sqrt{ \sum_{i=1}^{n} P_i^2} \cdot \sqrt{ \sum_{i=1}^{n} Q_i^2 }}$\\ \midrule
    Jaccard & $\frac{\sum_{i=1}^{n} (P_i - Q_i)^2}{(\sum_{i=1}^{n} P_i^2 + \sum_{i=1}^{n} Q_i^2 - \sum_{i=1}^{n} P_i \cdot Q_i)}$ \\ \midrule
    Dice & $\frac{\sum_{i=1}^{n} (P_i - Q_i)^2}{(\sum_{i=1}^{n} P_i^2 + \sum_{i=1}^{n} Q_i^2)}$\\ \midrule
    \multicolumn{2}{|c|}{Shannon's entropy family} \\ \midrule
    KL & $\sum_{i=1}^{n} P_i \cdot log_2  \frac{P_i}{Q_i} $ \\ \midrule
    Jeffreys & $\sum_{i=1}^{n} (P_i - Q_i)\cdot ln  \frac{P_i}{Q_i}$\\ \midrule
    JS & $\sqrt{\frac{KL(P,M) + KL(Q,M)}{2} }$ where $KL(\cdot,\cdot)$ is the KL function and\\
    	 &  $M$ is the pointwise mean vector of $P$ and $Q$\\ \midrule
    \multicolumn{2}{|c|}{$L_p$ Minkowski family} \\ \midrule
    Euclidean & $\sqrt{ \sum_{i=1}^{n} |P_i - Q_i|^2 }$\\ \midrule
    Cityblock & $\sum_{i=1}^{n} |P_i - Q_i|$\\ \midrule
    Chebyshev & $\mathop{max_{i=1}^{n}} |P_i - Q_i|$ \\ \midrule
    Minkowski (p=3) & $\sqrt[p]{ \sum_{i=1}^{n} |P_i - Q_i|^p }$\\ \midrule
    \multicolumn{2}{|c|}{$L_1$ family} \\ \midrule
    Braycurtis (S{\o}rensen) & $\frac{\sum_{i=1}^{n} |P_i - Q_i|}{ \sum_{i=1}^{n} (P_i + Q_i)}$ \\ \midrule
    Gower & $\frac{1}{n} \sum_{i=1}^{n} |P_i - Q_i|$ \\ \midrule
    Soergel & $\frac{\sum_{i=1}^{n} |P_i - Q_i|}{ \sum_{i=1}^{n} max(P_i,Q_i)}$ \\ \midrule
    Kulczynski & $\frac{\sum_{i=1}^{n} |P_i - Q_i|}{ \sum_{i=1}^{n} min(P_i,Q_i)} $ \\ \midrule
    Canberra & $\sum_{i=1}^{n} \frac{  |P_i - Q_i|}{  (P_i + Q_i)}$ \\ \midrule 
    Lorentzian & $\sum_{i=1}^{n} ln\ (1+ |P_i - Q_i|)$\\ \midrule
    \multicolumn{2}{|c|}{Fidelity or Squared-chord family} \\ \midrule
    Bhattacharyya & $-ln\ \sum_{i=1}^{n} \sqrt{P_i \cdot Q_i} $\\ \midrule
    Hellinger & $2 \sqrt{ 1 - \sum_{i=1}^{n} \sqrt{P_i \cdot Q_i} } $\\ \midrule
    Matusita & $\sqrt{ 2 - 2 \sum_{i=1}^{n} \sqrt{P_i \cdot Q_i} } $\\ \midrule
    Squaredchord  & $\sum_{i=1}^{n} ( \sqrt{P_i} - \sqrt{Q_i})^2 $\\ \midrule
    \multicolumn{2}{|c|}{Squared $L_2$ or $\chi^2$ family} \\ \midrule
    Pearson $\chi^2$ & $\sum_{i=1}^{n} \frac{(P_i - Q_i )^2}{Q_i}$\\ \midrule
    Neyman $\chi^2$ &  $\sum_{i=1}^{n} \frac{(P_i - Q_i )^2}{P_i}$\\ \midrule
    Squared $\chi^2$ & $\sum_{i=1}^{n} \frac{(P_i - Q_i )^2}{P_i + Q_i}$ \\ \midrule
    Prob. Symmetric $\chi^2$ & $2 \sum_{i=1}^{n} \frac{(P_i - Q_i )^2}{P_i + Q_i}$ \\ \midrule
    Divergence & $2 \sum_{i=1}^{n} \frac{(P_i - Q_i )^2}{(P_i + Q_i)^2}$\\ \midrule
    Clark & $\sqrt{\sum_{i=1}^{n} \Big( \frac{ |P_i - Q_i|}{P_i + Q_i} \Big)^2}$ \\ \midrule
    Additive Symmetric $\chi^2$ &  $\sum_{i=1}^{n} \frac{(P_i - Q_i )^2 \cdot (P_i + Q_i)}{ P_i \cdot Q_i}$\\
    \bottomrule
\caption{Dissimilarity metrics implemented in the \pkg{tegdet} library}
\label{tab:dissimilarityMetrics}
\end{longtable}


\section{Extensibility of the library}\label{app:extensibility}

The \pkg{tegdet} library has been designed for being extensible. In particular, the library can be extended to support new classes of dissimilarity metrics, in addition to the $28$ already implemented. The importance of implementing new metrics strives on augmenting the capabilities of the library to introduce new anomaly detector techniques. 
The extensibility has been achieved by applying a variant of the strategy design pattern~\citep{Gamma} to the software design of the library, see Figure~\ref{fig:tegdetCD} (Section~\ref{sec:impl}).

Since dissimilarity metrics define techniques used to compare graphs, then an abstract class, named \code{GraphComparator}, is proposed  
that will be specialized in as many classes as techniques want to be implemented. Figure~\ref{fig:tegdetCD} depicts three specialized classes as examples.

An abstract class owns one or more abstract methods, in this case \code{compare\_graphs} is the abstract method that needs to be implemented in each of the specialized classes, having the purpose of implementing the technique defined by the dissimilarity metric. 
Hence, a user of the library wanting to extend it with a new metric, say \code{<myMetric>}, only needs to create a new sub-class of \code{GraphComparator},
name it as \code{Graph<myMetric>Dissimilarity},  and implement the technique in the \code{compare\_graphs} method.

Finally, when a user wants to generate a TEG detector based on the new metric, she/he creates a new instance of the \code{TEGDetector} class
by setting the \code{metric} parameter to the name of the new metric in the \code{__init()__} instantiation method
---  see Figure~\ref{fig:modelConstruction}, message $1$  (Section~\ref{sec:impl}).


\section[Library repositories]{Library repositories} \label{app:repo}

The \pkg{tegdet} library is available  at the official \emph{PyPi} repository\footnote{PyPi url: \url{https://pypi.org/project/tegdet/}} as well as
at the \emph{GitHub} repository\footnote{GitHub url: \url{https://github.com/DiasporeUnizar/TEG}}.
In particular, the latter  also includes the following resources:
\begin{itemize}
	\item API documentation,
	\item Documentation about the installation and implementation,
	\item Dataset used in Section~\ref{sec:application},
	\item Test and example \proglang{Python} scripts.
\end{itemize}

\subsection{API of the library} \label{app:api}
Table~\ref{tab:TEGclassMethods} defines the API of the library, while Table~\ref{tab:TEGclassAttributes} details the attributes of the \code{TEGDetector} class, which are the parameters of the constructor of the API.

\begin{table}[h]
{\scriptsize
\begin{tabular}{p{7cm}p{8cm}}
\toprule
\textbf{{\bf tegdet} API methods} & \textbf{Description}\\\midrule
\code{init__(metric: string, n_bins: int =_N_BINS, n_obs_per_period: int =_N_OBS_PER_PERIOD, alpha: int =_ALPHA)}	&
 Constructor that initializes the TEGDetector input parameters (see Table~\ref{tab:TEGclassAttributes}).\\\midrule
\code{get_dataset(ds_path: string): DataFrame} 	&
 Loads the dataset from \code{ds_path}  file (comma-separated values format), renames the columns and returns it as a   
 \code{pandas Dataframe}.\\\midrule
 \code{build_model(training_dataset: Dataframe): ModelBuilder, float}	&
 Builds the prediction model based on the \code{training_dataset} and returns it (\code{ModelBuilder} object) 
 together with the time to build the model (\code{float} type).\\\midrule
 \code{predict(testing_dataset: Dataframe, model: ModelBuilder): numpy array of int, int, float} &
  Makes predictions on the \code{testing_dataset} using the \code{model}. It returns: the outliers (\code{numpy} array of \{0,1\} values) and total number of 
  observations (\code{int} type), and the time to make predictions (\code{float} type).\\\midrule
 \code{compute_confusion_matrix(ground_true: numpy array of int, predictions: numpy array of int): dict} &
  Computes the confusion matrix based on the ground true values and predicted values (\code{numpy} array of \{0,1\} values). 
  It returns the confusion matrix as a dictionary (\code{dict}) type. \\\midrule
 \code{print_metrics(detector: dict, testing_set: string, perf: dict, cm: dict)}		&  
Prints on the standard output: the \code{detector} (\code{dict} type including the metric and the input parameter setting), the \code{testing_set}, 
the performance metrics  \code{perf} (\code{dict} type including the time to build the model and the time to make predictions)  
and the confusion matrix \code{cm}.\\		\midrule
 \code{metrics_to_csv(detector: dict, testing_set: string, perf: dict, cm: dict, results_csv_path: string)}	& 
 Saves in the file with pathname \code{results_csv_path} (comma-separated values format):
 the \code{detector} (\code{dict} type), the \code{testing_set}, 
 the performance metrics  \code{perf} (\code{dict} type) and the confusion matrix  \code{cm} (\code{dict} type). \\\bottomrule
 \end{tabular} }
\caption{ {\bf tegdet} API (\texttt{TEGDetector} methods)}
\label{tab:TEGclassMethods}
\end{table}

\begin{table}[h]
{\scriptsize
\begin{tabular}{p{4cm}p{11cm}}
\toprule
 \textbf{Input parameters}       & \textbf{Description}   \\\midrule                                                                                    
 \code{__metric: string}   & Dissimilarity metric used to compare two graphs. \\                              
 \code{__n_bins: int}        & Level of discretization (number of levels).  Default value=  \code{__N_BINS (=30)}   \\
 \code{__n_obs_per_period: int} & Number of observation per period.  Default value= \code{__N_OBS_PER_PERIOD (=336)}\\
 \code{__alpha: int}         &  Significance level: \code{100-alpha}.  Default value=  \code{__ALPHA (=5)}   \\\bottomrule
 \end{tabular} }
\caption{\texttt{TEGDetector} attributes}
\label{tab:TEGclassAttributes}
\end{table}

\subsection{Library installation}\label{app:install}
The \pkg{tegdet} library has been implemented in \proglang{Python3} and it runs with Python versions $\geq 3.6.1$.
The library can be easily installed from the \emph{PyPi} repository as follows:
\begin{Code}
> pip3 install tegdet
\end{Code}

As an alternative, the \emph{GitHub} repository can be either cloned (or downloaded) and
the library can be installed from the distribution local to such repository using the command:
\begin{Code}
> pip3 install dist/tegdet-1.0.0-py3-none-any.whl
\end{Code}

Since the library depends on the \pkg{pandas}, \pkg{numpy} and \pkg{scipy}  Python packages, these 
should be installed before using the library.

The library dependencies are also listed in the \code{requirements.txt} file (available in the \emph{GitHub} repository), 
and  all the necessary packages  can be installed using the command:
\begin{Code}
> pip3 install -r requirements.txt
\end{Code}

\subsection{Example scripts} \label{app:scripts}
The \emph{GitHub} repository includes all the scripts and the dataset, which are necessary to reproduce the analysis carried out in Sections~\ref{sec:application} and~\ref{sec:guidelines_predictions}. 
In particular,  the \code{examples/energy} folder includes the following scripts:
\begin{itemize}
	\item \code{tegdet_variants.py}: builds and make predictions with TEG-detector variants (Sub-section~\ref{ss:variantsAnalysis}). 
	The code is shown in Listing~\ref{list:tegdet_variants};
	\item \code{tegdet_params_sensitivity.py}: carries out sensitivity analysis of TEG-detector parameters 
	(Section~\ref{sec:guidelines_predictions});
	\item \code{post_processing.py}: produces reports from the dataset (Sub-section~\ref{ss:dataset}) and the results of the analysis 
	(Sections~\ref{ss:basic_assessment} and~\ref{sec:guidelines_predictions}).
\end{itemize}
The first two scripts are examples of using the API, they both use the files in \code{dataset/energy} folder, 
and produce a results file \code{<name_of_the_script>_results.csv} in the  \code{script_results/ energy} folder.

The script  \code{post_processing.py} should be executed after the other two scripts.
Indeed,  it relies on both, the files in the \code{dataset/energy} folder and in the \code{script_results/energy} folder to produce reports 
(comparison of the testing datasets, performance and accuracy of the TEG-detectors). 
Since the script generates 3D plots, it is necessary to install the following package before running it:
\begin{Code}
> pip3 install matplotlib
\end{Code}
All the scripts can be run using the following command from the root directory of the repository{\footnote{The \code{PYTHONPATH} environment variable 
has to be set to the root directory of the \code{tegdet} project before running the scripts.}}:
\begin{Code}
> python3 examples/energy/<name_of_the_script>.py
\end{Code}

\paragraph{Reproducibility of the results.}
The execution time results presented in Sections~\ref{sec:application},~\ref{ss:basic_assessment} and~\ref{sec:guidelines_predictions} 
(i.e., time to build a model and time to make prediction) depend on the execution environment. 
Therefore, to reproduce such results, the script \code{post_processing.py} includes the following lines:
\begin{lstlisting}[basicstyle=\ttfamily\tiny,numbers=none,backgroundcolor = \color{vlgray}]
TEGDET_VARIANTS_RESULTS_PATH = "/script_results/energy/tegdet_variants_results_ref.csv"
TEGDET_PARAMS_SENSITIVITY_RESULTS_PATH = "/script_results/energy/tegdet_params_sensitivity_results_ref.csv"
#Uncomment these lines to analyse the results once the scripts have been run
#TEGDET_VARIANTS_RESULTS_PATH = "/script_results/energy/tegdet_variants_results.csv"
#TEGDET_PARAMS_SENSITIVITY_RESULTS_PATH = "/script_results/energy/tegdet_params_sensitivity_results.csv"
\end{lstlisting}
The first two lines set the result paths to the files generated during the analysis described in Sections~\ref{sec:application}, ~\ref{ss:basic_assessment} and~\ref{sec:guidelines_predictions}, by the scripts \code{tegdet_variants.py} and \code{tegdet_params_sensitivity.py}, respectively. 
The last two lines set the result paths to the files that will be  generated by a new execution of the above scripts.

\begin{lstlisting}[caption={Script \texttt{tegdet\_variants.py}}, label={list:tegdet_variants}]
import os
import pandas as pd
import numpy as np
from tegdet.teg import TEGDetector

#Input datasets/output results paths
TRAINING_DS_PATH = "/dataset/energy/training.csv"
TEST_DS_PATH = "/dataset/energy/test_"
RESULTS_PATH = "/script_results/energy/tegdet_variants_results.csv"

#List of testing
list_of_testing = ("normal", "anomalous")

#List of metrics (detector variants)
list_of_metrics = ("Hamming", "Cosine", "Jaccard", "Dice", "KL", "Jeffreys", "JS", 
                    "Euclidean", "Cityblock", "Chebyshev", "Minkowski", "Braycurtis",
                    "Gower", "Soergel", "Kulczynski", "Canberra", "Lorentzian",
                    "Bhattacharyya", "Hellinger", "Matusita", "Squaredchord",
                    "Pearson", "Neyman", "Squared", "Probsymmetric", "Divergence",
                    "Clark", "Additivesymmetric" )

#Parameters: default values
n_bins = 30
n_obs_per_period = 336
alpha = 5

def build_and_predict(metric):
    cwd = os.getcwd() 
    train_path = cwd + TRAINING_DS_PATH

    tegd = TEGDetector(metric)
    #Load training dataset
    train = tegd.get_dataset(train_path)
    #Build model
    model, time2build = tegd.build_model(train)

    for testing in list_of_testing:

        #Path of the testing
        test_path = cwd + TEST_DS_PATH + testing + ".csv"               
        #Load testing dataset
        test = tegd.get_dataset(test_path)
        #Make prediction
        outliers, n_periods, time2predict = tegd.predict(test, model)
        #Set ground true values
        if testing == "anomalous":
            ground_true = np.ones(n_periods)        
        else:
            ground_true = np.zeros(n_periods)

        #Compute confusion matrix
        cm = tegd.compute_confusion_matrix(ground_true, outliers)

        #Collect detector configuration
        detector = {'metric': metric, 'n_bins': n_bins, 
        		'n_obs_per_period':n_obs_per_period, 'alpha': alpha}
        #Collect performance metrics in a dictionary
        perf = {'tmc': time2build, 'tmp': time2predict}

        #Print and store basic metrics
        tegd.print_metrics(detector, testing, perf, cm)
        results_path = cwd + RESULTS_PATH
        tegd.metrics_to_csv(detector, testing, perf, cm, results_path)
        
if __name__ == '__main__':

    for metric in list_of_metrics:
        build_and_predict(metric)
\end{lstlisting}

\end{appendix}


\end{document}